\documentclass[Afour,sageh,times]{sagej}

\usepackage{moreverb,url}

\usepackage[colorlinks,bookmarksopen,bookmarksnumbered,citecolor=red,urlcolor=red]{hyperref}

\usepackage{graphics} 
\usepackage{epsfig} 
\usepackage{mathptmx} 
\usepackage{amsmath} 
\usepackage[utf8]{inputenc}
\usepackage[T1]{fontenc}
\usepackage{mathtools}
\usepackage{enumitem}
\usepackage{xcolor}
\usepackage{float}
\usepackage{gensymb}
\usepackage{bm}
\usepackage{autobreak}
\usepackage{svg}
\usepackage{mathrsfs}
\usepackage{tabularx}
\usepackage{array, makecell}
\usepackage{graphicx}
\usepackage{cellspace}
\usepackage{textcomp}
\usepackage{hyperref}
\usepackage{cleveref}
\usepackage{booktabs}
\usepackage{algorithm}
\usepackage{algpseudocode}
\usepackage{blkarray}
\usepackage{leftindex}

\usepackage[title]{appendix}

\usepackage{amssymb}

\usepackage{xcolor}

\renewcommand{\Re}{\mathbb{R}}

\newtheorem{theorem}{Theorem}
\newtheorem{lemma}[theorem]{Lemma}

\newcommand{\corrlab}[2]{{{#2}}}

\newcommand{\cusst}[1]{{}}

\DeclareMathAlphabet{\mathcal}{OMS}{cmsy}{m}{n}  
\algnewcommand{\LineComment}[1]{\State \(\triangleright\) #1}

\newcommand\BibTeX{{\rmfamily B\kern-.05em \textsc{i\kern-.025em b}\kern-.08em
T\kern-.1667em\lower.7ex\hbox{E}\kern-.125emX}}

\begin{document}

\runninghead{Shen \textit{et~al.}}

\title{Modeling the frictional contact of continuum robots with environments through linear complementarity problem}

\author{Jia Shen\affilnum{1,$^{*}$}, Junhyoung Ha\affilnum{{1,2,$^{*}$}} and Yue Chen\affilnum{1}}


\affiliation{
\affilnum{1}The Wallace H. Coulter Department of Biomedical Engineering, Georgia Institute of Technology/Emory University, Atlanta, GA, USA.\\
\affilnum{2}Department of Mechanical Engineering, Ulsan National Institute of Science and Technology, Ulsan, South Korea.
\affilnum{$^{*}$}These authors contributed equally to this work.
}

\corrauth{
Yue Chen, The Wallace H. Coulter Department of Biomedical Engineering, Georgia Institute of Technology/Emory University, 313 Ferst Dr NW, Atlanta, GA 30332, USA.
}


\email{yue.chen@bme.gatech.edu}

\begin{abstract}
Accurate friction modeling in continuum robots is essential for precise motion control and path planning. However, current approaches make predictions only for the mechanism's internal friction that is limited by the boundary forces of the Coulomb friction cone, which fail to characterize the static friction and motion hysteresis. In this paper, we propose a novel friction model based on complementarity constraints under the Coulomb friction assumption.
This model predicts various friction effects caused by environmental interactions, including both static and sliding frictions and hysteresis behavior, as well as the corresponding robot deformation induced by these interactions, which are crucial for the practical motion planning and control of continuum robots. Additionally, real-time implementation of this model can be achieved, meeting the demand for computational efficiency in robotic applications. Experimental validations demonstrated that this model is able to reduce the maximum tip position error from $45.4$ mm in the frictionless model to $5.85$ mm in our friction-included model with a $287.9$ mm-long tendon-driven continuum robot.
\corrlab{R1-5}{
The source code for the proposed model is publicly available at \url{https://github.com/Jia0Shen/LCP-Continuum}.
}
\end{abstract}


\maketitle

\footnotetext[0]{*Accepted for publication in the
  \emph{International Journal of Robotics Research}, July 2026.}
  
\setcounter{secnumdepth}{0}
\section{Introduction}

Continuum robots, characterized by their flexible structure and enhanced dexterity, have gained considerable attention in various fields including minimally invasive surgery~\citep{Review_continuum2015, Review_continuum2023}, robotic rescue~\citep{ContinuumRescue}, and inspections~\citep{ContinuumInsppection}. Unlike traditional rigid-body robots, continuum robots can navigate in limited workspaces efficiently and safely due to their inherent compliance and adaptability.
However, it poses significant challenges to control continuum robots when they encounter internal and external frictional contacts, due to the degraded accuracy of available frictionless models.
For instance, \citet{ctr_friction_ha2019} demonstrated that incorporating friction in a concentric tube robot (CTR) model can lead to centimeter-level tip position errors compared to frictionless predictions. 
Similarly, \citet{TDCR_friction_Gao} showed that friction in tendon-driven continuum robots (TDCRs) can induce large angular deviations, exceeding tens of degrees, under typical actuation conditions.


Accurate modeling of continuum robots requires consideration of multiple types of frictional contacts, including internal and environmental contacts. Existing models primarily address internal friction, such as inter-tube friction in CTRs or tendon friction in tendon-driven robots \citep{shen2026ral}, while environmental friction modeling for continuum robots has been largely unexplored. Notable prior works include \citet{Nabil_friction2024, TDCR_stability2023, ctr_friction_ha2019, frictionCTR2011, TDCR_friction_Gao, TDCR_friction_2015tackling}, which have highlighted the effects of frictional direction for CTRs and TDCRs, respectively. However, these models consider only sliding (kinetic) friction under the simplifying assumption that the relative sliding direction at each contact is determined by the actuation direction. This approach, while capturing the maximum friction effect, leads to inaccurate predictions of sudden friction force changes and corresponding robot configurations when the actuation direction changes. 

\begin{figure}[t!] 
    \centering
    \includegraphics[width = 0.9\linewidth]{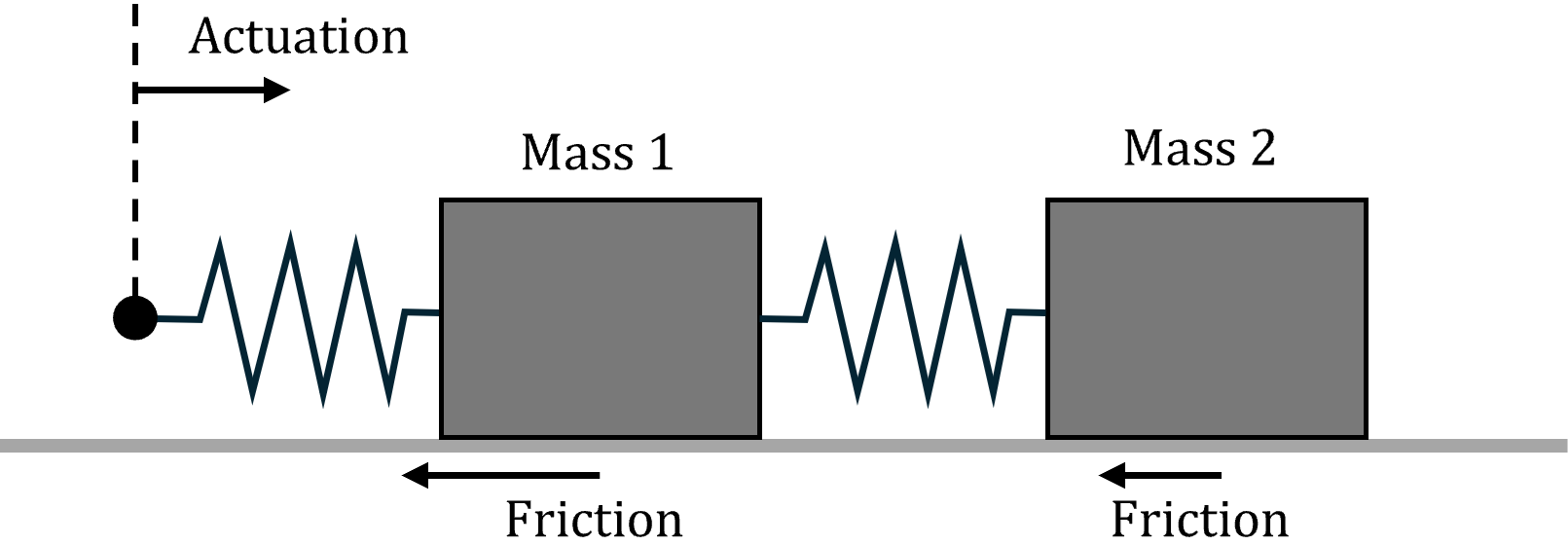}
    \caption{A serial mass-spring system subjected to ground friction. The base displacement does not always cause immediate movements of the masses or changes in friction forces.}
    \label{fig:spring_mass}
\end{figure}

In reality, due to the compliant nature of continuum robots, neither the friction force nor the sliding direction changes immediately by actuation. This can be intuitively explained via a serial mass-spring system with the consideration of friction force in Fig.~\ref{fig:spring_mass}.
The friction force of the first mass changes immediately with the base displacement, yet a finite displacement is required to initiate its movement. In contrast, the friction force on the second mass remains unchanged until the first mass moves. Actuating the base displacement in varying directions further complicates this behavior, highlighting the challenge that has led previous friction models to rely on the sliding friction assumption.

Predicting the continuous transition of the friction forces and resulting robot configuration is complex due to their hysteretic nature, that is, they depend on the actuation history. The primary modeling challenge lies in the fact that the friction force at a given time instant may take any value within the friction cone, and that determining the contact location is inherently coupled with determining the friction force. 
To tackle these challenges, the robot configuration, contact locations, and friction forces must be solved simultaneously over an actuation history.


We draw inspiration from the field of rigid robot mechanics. 
In multi-body dynamics, frictional contact is commonly handled using impulse-based time-stepping methods, in which contact forces and moments are represented as impulses applied over discrete time intervals \citep{MLS_robotics_txtbk,LCP1997_A_P}.
These impulses cause instantaneous changes in rigid body velocities, constructing a linear equation between the friction forces and velocity changes. Subsequently, the friction forces and velocity changes are solved simultaneously through a linear complementarity problem (LCP) subject to a set of complementarity constraints (CCs) that capture the physics of frictional contacts~\citep{LCP1997_A_P}.
Although such complementarity-based contact modeling has also been applied to continuum robots \citep{TRO2024_LCP, faure2012sofa}, the resulting formulations are typically dynamics-based and are therefore not well suited to the quasi-static operating regime of continuum robots. Incorporating dynamics increases model complexity and may also make the formulation more sensitive to modeling uncertainty, particularly for lightweight robots whose inertial effects are small \citep{bajo2016hybrid}.

In this paper, we present a novel LCP-based friction model for environmental contacts in continuum robots. By integrating complementarity constraints with the Cosserat rod theory ~\citep{rucker2010kinematics, rucker2011statics}, the proposed model enables consistent prediction of frictional contact forces and resulting robot configurations.
%
%
The main contributions of our paper are summarized as follows:
\begin{itemize}
    \item We reformulate the conventional mass/inertia-dependent LCP into a compliance-based formulation for continuum robots, enabling efficient computation while capturing static friction and hysteresis behaviors.
    \item We develop an iterative algorithm that simultaneously detects environmental contacts, computes the associated contact forces, and updates the robot's shape.
    \item We provide mathematical proof of the solution existence of the LCP-based method and its equivalence to the frictionless model in special scenarios, bridging the gap between our method and other existing models.
\end{itemize}

The rest of this paper is formulated in the following structure. \hyperref[sec:related_works]{Related work} section summarizes the existing friction models of continuum robots, followed by the \hyperref[sec:preliminaries]{Preliminaries}. The mathematical modeling of the LCP-based friction model and the contact detection algorithm used in the model are developed in \hyperref[sec:method_general]{External friction modeling of continuum robots} section. The proposed approach is then applied to an elastic rod in \hyperref[sec:rod]{Case study: Application to an elastic rod}. The proposed model is validated in the section of \hyperref[sec:simu]{Numerical simulations} and \hyperref[sec:experiments]{Experimental results}, followed by the \hyperref[sec:conclusion]{Conclusion}.

\section{Related work} \label{sec:related_works}


Friction modeling has been extensively studied for decades in the field of rigid robot dynamics, leading to the development of a standard approach. However, the continuum robotics field has witnessed a relatively shorter history in studying friction modeling. In this section, we will summarize the related work in both fields.

\subsection{Friction Model for Continuum Robots}

Existing friction models for continuum robots explored the mechanism's internal friction. To model the inter-tube friction in CTRs, \citet{frictionCTR2011} considered a tube pair with identical curvatures in unidirectional rotation. The tip torsional friction was considered to model the additional tube twist induced by friction, and a prediction on the base rotation vs. tip rotation was presented and compared with experimental outcomes. In \citet{ctr_friction_ha2019}, once again, the torsional friction was considered for general CTRs along with the contact location prediction. This approach assumed sliding friction forces, i.e., the maximum Coulomb friction force, providing only the bounding effect of friction in robot configuration.

The friction in tendon-driven continuum robots (TDCR) has been modeled to account for the loss of tension along the tendon and the hysteresis effects \citep{TDCR_friction_2015tackling, TDCR_friction_Gao, TDCR_stability2023}. An analytical model of distributed tendon friction for TDCR was proposed in \citet{Yang2022_DistributeFriction}, solving the robot's deformation and the unknown friction force iteratively. \citet{TDCR_friction_state} proposed a friction model based on a state machine to predict the sticking/sliding state change along the tendon, and the static friction was neglected at the sticking state. In \citet{Gao2017_tendonFriction_static,Gao2020_tendonFriction_sensing}, the static friction was approximated using a constant COF smaller than the sliding COF, and the value was obtained through experimental calibrations. All these models predicted the Coulomb friction force by $\mathbf{f} = - \mu f_\text{n} \mathbf{v} / \lVert \mathbf{v} \rVert$ with a constant COF ($\mu$) under a simplifying assumption of known relative sliding direction ($\mathbf{v}$) and normal contact force ($f_\text{n}$), which failed to capture the continuous change of friction force within the Coulomb friction cone.

\subsection{Contact Model for Rigid Robots}

In rigid-body robotics, contact has been extensively studied in the context of forward dynamics, both for realistic simulation and for model-based control \citep{Posa_2014Direct, Alp2022_contactMPC}. A common formulation enforces normal contact through a complementarity constraint (CC) between the normal relative velocity and the normal contact force~\citep{trinkle1997dynamic}. Within impulse-based time-stepping methods, the forward dynamics can be expressed as a linear force--velocity relation. However, Coulomb friction is inherently conic and nonlinear, so incorporating friction into this framework leads, in general, to a nonlinear complementarity problem (NCP). \citet{LCP1997_A_P} addressed this issue by approximating the friction cone with a polyhedral representation, thereby expressing friction through additional CCs and linear equations and recovering a linear complementarity problem (LCP). The existence of solutions to the resulting LCP has been established in~\citet{LCP1997_A_P, LCP_Intro}.

More recent rigid-body physics simulators implement contact using different numerical strategies. Bullet \citep{pybullet} relies on iterative constraint solvers (e.g., sequential impulse / projected Gauss--Seidel) to approximate complementarity conditions, while MuJoCo \citep{MuJoCo} adopts a smooth penalty-based formulation for contact and friction. Drake \citep{drake} supports point and hydroelastic contact models, and NVIDIA Isaac Gym \citep{IsaacGym} uses GPU-accelerated PhysX simulation with a Temporal Gauss--Seidel solver for large-scale parallel robot learning. 
Despite these differences, these simulators are primarily designed for dynamic rigid-body systems and do not incorporate the Cosserat rod model, which limits their direct applicability to soft and continuum robots. \citet{TRO_simulator_review} provides a comprehensive review of the recent contact models.

\subsection{Contact Models for Soft and Continuum Robots 
}

Contact modeling for soft and continuum robots has been approached through both general simulation frameworks and Cosserat-rod-based models. \citet{Duriez2006} addressed contact and friction between deformable objects using a complementarity-based formulation for real-time haptic simulation. PyElastica \citep{PyElastica} provides a discrete Cosserat rod framework with compliant contact modules, in which contact is enforced through spring--damper-type penalty terms; however, such penalty-based treatment smooths the contact response and may introduce modeling error. SOFA \citep{faure2012sofa} provides a high-fidelity FEM-based simulation environment for biomechanics and robotics that supports constraint-based contact solvers, including an LCP solver \citep{SOFA_LCP}. However, its FEM formulation leads to high-dimensional problems and increased computational cost.

Prior Cosserat rod-based models have also incorporated complementarity formulations for environmental contact \citep{LCP_noFriction2019}, although friction was not considered. \corrlab{R2-1}{More recently, \citet{TRO2024_LCP} proposed a dynamic Cosserat rod formulation for soft slender robots with frictional environmental contact, in which the contact and friction conditions were formulated as nonlinear complementarity constraints and smoothed into equality constraints for real-time optimization-based simulation. } Similar to rigid-body contact models \citep{LCP1997_A_P}, these approaches use inertial quantities to relate contact forces to deformation. In many continuum-robot applications, however, such dynamic modeling is not essential, since these systems are typically operated in a quasi-static regime \citep{Review_continuum2015}, and abrupt motions such as snapping are intentionally avoided in medical settings for safety \citep{review2022Dupont}. 
Moreover, introducing dynamics increases model complexity and may exacerbate sensitivity to modeling uncertainty, particularly for slender robots with small mass and poorly characterized inertial properties \citep{bajo2016hybrid}. This motivates a quasi-static reformulation in which contact is expressed directly through compliance rather than through dynamics.

\section{Preliminaries} \label{sec:preliminaries}

This section will review the continuum robot kinematics, the CC-based friction formulation, and the LCP, which will form the basis of our friction model. Table \ref{notation} summarizes the notation used throughout this paper.

\subsection{Kinematics Model of Continuum Robots} \label{sec:review_model}

\begin{table}[t!]
\renewcommand{\arraystretch}{1.2}
\caption{Nomenclature}
\label{notation}
\small\sf\centering
\begin{tabularx}{8.7cm}{l X}
\toprule
 Notation & Definitions\\
\midrule
 $s$ &  Arc-length parameter for the central axis. \\
 $n$ &  Number of arc-length discretization points.\\
 $\mathbf{p}(s)$  & Position vector of the material frame at $s$.\\
 $\mathbf{R}(s)$  & Rotation matrix of the material frame at $s$. \\
 $\mathbf{T}(s)$  & Homogeneous rigid-body transformation matrix of the material frame at $s$.\\
 $\mathbf{u}(s)$  & {Curvature vector of the continuum robot at $s$.}\\
 $\mathbf{\hat{u}}(s)$  & {Precurvature vector of the continuum robot at $s$.}\\
 $\mathbf{\underline{p}}$, $\mathbf{\underline{u}}$, $\mathbf{\underline{\hat{u}}}$   & Vectors comprised of all discretized $\mathbf{p}(s)$, $\mathbf{u}(s)$, and $\mathbf{\hat{u}}(s)$. \\
 $\mathbf{K}$  & $3\times 3$ stiffness matrix, defined as $\mathbf{K} = \text{diag}([K_b, K_b, K_t])$, where $K_b$ and $K_t$ are bending and torsional stiffnesses. \\
 $\mathbf{\underline{K}}$  & {$3n\times 3n$ matrix comprised of $\mathbf{K}$ corresponding to $\mathbf{\underline{u}}$. }\\
 $\Phi(\mathbf{p})$ & Signed distance from a point $\mathbf{p}$ to the environment. \\
 $\mathbf{D}$ & Matrix whose columns represent the tangent directions of the polygonized friction cone. \\
 $\bm{\beta}$ & Vector comprised of magnitudes along the tangent directions in $\mathbf{D}$. \\
 $\mathbf{q}$ & Actuation vector of the continuum robot. \\
 $\mathbf{w}$ & Contact wrench applied to the continuum robot. \\
 $\mathbf{f}$ & Contact force applied to the continuum robot.\\
 $\mathbf{v}$ & Relative displacement between the robot and the environment at the contact point.\\
 $\mathbf{n}$ & Contact normal vector. \\
 $\nu$ & Poisson's ratio. 
 \\
 $\mu$  & Coefficient of friction (COF). \\
 $(\cdot)^{\prime}$ &  Rate-of-change w.r.t. arc-length $s$, i.e. $\frac{\partial}{\partial s}(\cdot)$. \\
 $[\cdot]_{\times}$ & Element of the Lie algebra $\text{so}(3)$ or $\text{se}(3)$ in $3 \times 3$ or $4 \times 4$ matrix representation~\citep{lynch2017modern}. \\
 $\perp$ & Notation of complementarity constraints: $0\leq a \perp b \geq 0$ is equivalent to $a \geq 0$, $b \geq 0$, and $ab=0$. \\
\bottomrule
\end{tabularx}
\end{table}

Cosserat rod theory has been widely used to represent the geometry of continuum robots \citep{rucker2010kinematics, rucker2011statics}, which assigns a material frame along the center-line curve such that the $z$-axis of the frame is tangent to the curve. The homogeneous rigid-body transformation $\mathbf{T}(s) \in \text{SE}(3)$ is used to describe the evolution of the position and orientation of the frame:
\begin{equation}
    \mathbf{T}(s) =
    \begin{bmatrix}
         \mathbf{R}(s) & \mathbf{p}(s) \\
        \mathbf{0}_{1\times3} & 1 
    \end{bmatrix},
\end{equation}
where $\mathbf{p}(s) \in \Re^3$ and $\mathbf{R}(s) \in \text{SO}(3)$ are the position and orientation, respectively, and $s \in \Re$ is the arc-length parameter.

The rod experiences different modes of elastic deformations under external loads, including bending and torsion. These effects are related to the rate-of-change of $\mathbf{T}(s)$ by:
\begin{equation} \label{eq:p_u_relation}
    \mathbf{T}^{\prime}(s) = \mathbf{T}(s) \left[ \bm{\xi}(s) \right]_{\times},
\end{equation}
where $\bm{\xi}(s) \in \Re^{6}$ is the body twist of the material frame, 
$[\cdot]_{\times}$ denotes the operator that maps the twist to an element in $se(3)$ ~\citep{lynch2017modern},
and $(\cdot)'$ denotes the arc-length derivative.
\corrlab{R1-1}{
The twist can be expressed as $\bm{\xi}(s) = [\mathbf{u}(s)^T, \bm{\nu}(s)^T]^T$, where $\mathbf{u}(s) \in \Re^3$ is the curvature vector, and $\bm{\nu}(s) \in \Re^3$ is the rate of change of position. Under the assumption of a shear-free and inextensible Cosserat rod \citep{rucker2010kinematics}, $\bm{\nu}(s)$ is the $z$-directional unit vector $\mathbf{e}_z \in \Re^3$.
}

Combining (\ref{eq:p_u_relation}) with the continuum mechanics, models of continuum robots with actuation variable $\mathbf{q}$ and a point wrench $\mathbf{w}$ can be written as a set of ordinary differential equations (ODEs):
\begin{equation} \label{eq:ode_cosserat}
\begin{aligned}
    \mathbf{T}^{\prime}(s) & = \mathbf{T}(s) \left[ \bm{\xi}(s) \right]_{\times}, \\
    \mathbf{y}^{\prime}(s) & = g(s,\mathbf{y}, \mathbf{q}, \mathbf{w}),
\end{aligned}
\end{equation}
where $g(\cdot)$ is the system function of the Cosserat rod model, and $\mathbf{y}(s)$ is the state vector of a specific continuum robot formulation.
\corrlab{R1-4}{For instance, the state vector of a single elastic tube can be written as $\mathbf{y}(s)=\mathbf{u}(s)$. }

The transformation at a discretized arc-length location $\mathbf{T}(s_i)$ can be integrated through $\mathbf{T}(s_{i}) = \mathbf{T}(s_{i-1}) \exp{(\left[ \bm{\xi}(s_{i-1}) \right]_{\times}(s_{i}-s_{i-1}))}$ to preserve the SE(3) structure.
To solve the ODEs in (\ref{eq:ode_cosserat}), a boundary value problem (BVP) is formed given the geometric constraints at the robot base and the force/moment equilibrium at the tip. 
We denote $\mathcal{F}(\cdot)$ as the robot's configuration solved from the BVP of the Cosserat rod model: 
\begin{equation} \label{eq:F_coss}
    \mathbf{T}(s) = \mathcal{F}(\mathbf{q}, \mathbf{w}).
\end{equation}
%
\subsection{Review of CC-based Coulomb Friction Modeling}
\label{sec:preliminary_CC}
The frictional contact between a rigid object and a rigid surface has been modeled through a set of CCs \citep{LCP1997_A_P}, which established the relationship between the incremental displacement at the contact point over a time step, $\mathbf{v}$, and the contact force, $\mathbf{f}$.

Suppose that the object is in contact with the surface. We first focus on the CC in the contact normal direction. The incremental displacement and contact force along the normal direction are defined as
\begin{equation} \label{eq:def_vn_fnorm}
    v_\text{n} = \mathbf{n}^T \mathbf{v}, \quad  f_\text{n} = \mathbf{n}^T \mathbf{f},
\end{equation}
where $\mathbf{n}$ is the contact normal vector. To avoid penetration, the normal incremental displacement must be non-negative, i.e. $v_\text{n} \geq 0$, with a strict inequality indicating take-off (contact breaks). Additionally, the surface only exerts a pushing force on the object; thus $f_\text{n} \geq 0$ holds.

In the case of take-off, the object is lifted up ($v_\text{n} > 0$), resulting in zero contact force ($f_\text{n} = 0$), while contact must not break ($v_\text{n} = 0$) for a non-zero contact force ($f_\text{n} > 0$). Both cases can be incorporated in a single complementarity condition, known as the Signorini contact condition:
\begin{equation} \label{eq:CC_origin}
    v_\text{n} \geq 0,\quad f_\text{n} \geq 0, \quad v_\text{n} \cdot f_\text{n} = 0,
\end{equation}
or equivalently, expressed as
\begin{equation} \label{eq:CC_normal}
    0 \leq v_\text{n} \perp f_\text{n} \geq 0.
\end{equation}

Subsequently, friction is studied in the tangent direction of the surface. Following Coulomb's friction law, the contact force lies inside the friction cone that can be approximated as a polygon with $m$ sides, as shown in Fig. \ref{fig:delta_p}(a), which leads to a decomposition of $\mathbf{f}$:
\begin{equation} \label{eq:f_decompose}
    \mathbf{f}  = f_\text{n} \mathbf{n} + \mathbf{D} \bm{\beta},
\end{equation}
where $\mathbf{D} = [\mathbf{d}_1, \ldots, \mathbf{d}_m]$ with $\mathbf{d}_k$ being a unit vector representing the $k$-th tangent direction, $\bm{\beta}=[\beta_1, \ldots, \beta_m]^T$ with $\beta_k$ being the magnitude of the friction force along $\mathbf{d}_k$. Then the inequality of Coulomb's friction law can be expressed as 
\begin{equation} \label{eq:f_mu_n}
    \eta \equiv \mu f_\text{n} - \mathbf{e}^T \bm{\beta} \geq 0,
\end{equation}
where $\mu$ denotes the COF, and $\mathbf{e}=[1,\ldots,1]^T \in \Re^m$. The value of the friction force $\bm{\beta}$ depends on the magnitude of the incremental displacement in tangent direction. 
For the stick phase ($\mathbf{D}^T \mathbf{v} = \mathbf{0}$), the friction force is not uniquely determined by Coulomb’s friction law and lies within the friction cone ($\eta > 0$). In contrast, the friction force lies on the boundaries of the friction cone ($\eta=0$) when relative sliding occurs ($\mathbf{D}^T \mathbf{v} \neq \mathbf{0}$). 
The magnitude of the incremental displacement can be represented by a slack variable $\lambda$ to formulate a similar linear complementarity condition:
\begin{subequations} \label{eq:CC_tangent}
\begin{align}
        0 \leq \bm{\sigma} ~ & \perp ~ \bm{\beta} \geq 0, \label{eq:CCs_friction1} \\
        0 \leq \eta  ~ & \perp ~ \lambda \geq 0,\label{eq:CCs_friction2}
\end{align}
\end{subequations}
where $\bm{\sigma} \equiv \lambda \mathbf{e} + \mathbf{D}^T \mathbf{v}$.
We refer readers to \citet{LCP1997_A_P} for a detailed derivation of (\ref{eq:CC_tangent}). 


The final CC at the contact point is summarized by combining (\ref{eq:CC_normal}) and (\ref{eq:CC_tangent}):
\begin{equation} \label{eq:CCs}
    0 \leq \begin{bmatrix}
        v_\text{n} \\ \bm{\sigma} \\ \eta
    \end{bmatrix} \perp 
    \begin{bmatrix}
        f_\text{n} \\ \bm{\beta} \\ \lambda
    \end{bmatrix} \geq 0
\end{equation}
with $\bm{\sigma} = \lambda \mathbf{e} + \mathbf{D}^T \mathbf{v} $, $\eta = \mu f_\text{n} - \mathbf{e}^T \bm{\beta}$, and $\lambda$ as a slack variable.

\subsection{Review of LCP} \label{sec:lcp_review}

LCP is a standard approach to solving for unknowns in CCs subject to linear equations.
Given parameters $\mathbf{M} \in \Re^{n \times n}$ and $\mathbf{h} \in \Re^{n}$, an LCP is defined as a constraint satisfaction problem:
\begin{equation} \label{eq:lcp_preliminary}
    \begin{aligned}
        \text{Find} & ~ \mathbf{x}, \mathbf{z} \in \Re^{n}, \\
        \text{Subject to} & ~ \mathbf{z} = \mathbf{M} \mathbf{x} + \mathbf{h}, \\
        & 0 \leq \mathbf{x} \perp \mathbf{z} \geq 0.
    \end{aligned}
\end{equation}
The solution set of (\ref{eq:lcp_preliminary}) is denoted as $\text{LCP}(\mathbf{M}, \mathbf{h})$, and the solution existence can be shown by the following lemma:
\begin{lemma}
$\text{LCP}(\mathbf{M}, \mathbf{h})$ is non-empty if the following conditions hold:
\begin{enumerate}
    \item $\mathbf{M}$ is copositive, i.e. $\mathbf{z}^T \mathbf{M} \mathbf{z} \geq 0$ for $\mathbf{z} \geq \mathbf{0}$.
    \item If $\mathbf{z} \in \text{LCP}(\mathbf{M}, \mathbf{0})$, then $\mathbf{h}^T \mathbf{z} \geq 0$.
\end{enumerate}
\end{lemma}
Once the LCP is proved solvable, Lemke's algorithm can be used to find the solution, with a maximum computational cost of order $2^n$. We refer the readers to \citet{LCP_Intro} for detailed descriptions of the LCP.

\section{External Friction Modeling of Continuum Robots} \label{sec:method_general}

This section presents a novel kinematics model of continuum robots that incorporates external friction forces through CCs. The proposed CC-based approach solves friction through motion and requires an intricate contact detection process. The algorithm for implementation is detailed at the end of this section. 

\subsection{LCP Formulation of External Friction} \label{sec:lcp_general}

We will derive an LCP formulation to compute the robot states incrementally through discrete time steps, considering a moving environment with known movement. The assumptions made in our formulations are as follows:
\begin{itemize}
    \item[(i)] Spatial linear approximation: The contact surface is locally approximated to be planar;
    \item[(ii)] Temporal linear approximation: Linearized kinematics (i.e., differential kinematics) is used to describe the robot's motion over a small time step.
    \item[(iii)] Hard contacts: The environment can be modeled as rigid bodies that exert supporting force along the normal direction.
    \item[(iv)] Known geometry: The environment's geometry is obtained beforehand or through real-time sensing.
\end{itemize}

\subsubsection{Differential Kinematics}
\label{subsubsec:diff_kin}
\begin{figure} 
    \centering
    \includegraphics[width = \linewidth]{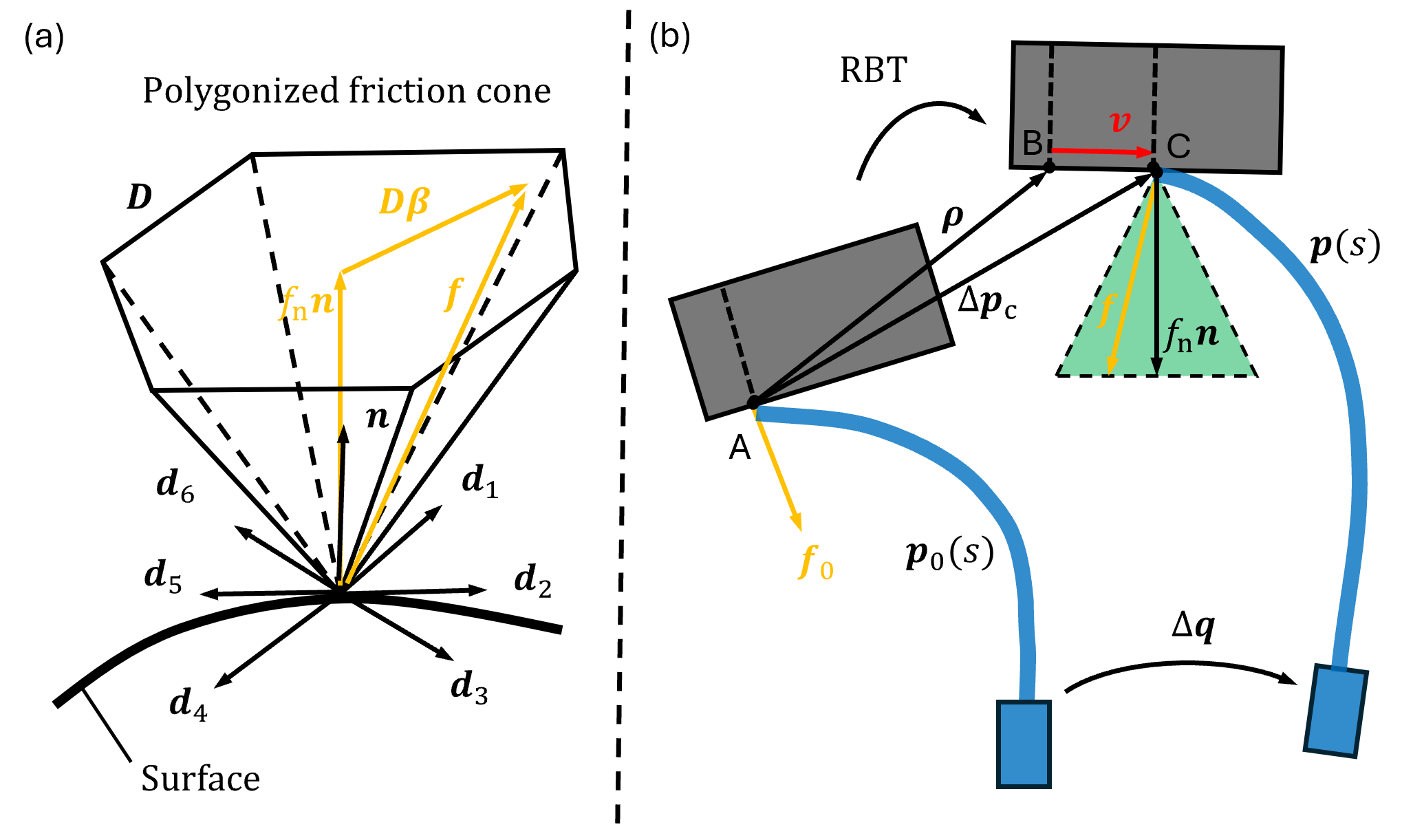}
    \caption{Geometric representation of the frictional contact between a continuum robot and rigid environment. (a) The polygonized Coulomb friction cone. (b) Shape change of the robot over a time interval. A rigid-body transformation is performed to the obstacles in the environment and the robot is actuated by $\Delta \mathbf{q}$. The contact force ($\mathbf{f}$) inside the Coulomb friction cone and the relative displacement ($\mathbf{v}$) are solved simultaneously through an LCP. $\bm{\rho}$ and $\Delta \mathbf{p}_c$ denote the vector of the environment movement and displacement of contact points, respectively.
    }
    \label{fig:delta_p}
\end{figure}

Consider a continuum robot with $d$ degrees of freedom (DoF) interacting with the environment through a total number of $p$ contacts, as depicted in Fig. \ref{fig:delta_p}(b).
The differential kinematics captures how these contact points move due to robot actuation and contact force changes, which provides a linear equation to complete the LCP formulation for friction modeling.

Let us define the collection vectors of contact points and wrenches as
\begin{equation} \label{eq:def_pc_wc}
\mathbf{p}_c = (\mathbf{p}^1, \ldots,  \mathbf{p}^p) \in \Re^{3p}, \
\mathbf{w} = (\mathbf{w}^1, \ldots,  \mathbf{w}^p) \in \Re^{6p}.
\end{equation}
Here, the wrenches are expressed at the robot's local frame $\mathbf{T}(s)$. 
We use $\mathbf{q}$ to denote the generalized coordinates of the robot. 
In our formulation, these coordinates correspond directly to the actuator variables (e.g., tube insertions and rotations) under the Cosserat rod model. 
As the robot is $d$-DoF, the actuation vector can be expressed as $\mathbf{q}=[q_1,\ldots,q_{d}]$. The motion of the $i$-th contact point on the continuum robot can be described w.r.t. the actuation change $\dot{\mathbf{q}}$ and wrench change $\dot{\mathbf{w}}$ \citep{rucker2011_Jacob}.
\begin{equation} \label{eq:T_J_start}
    (\mathbf{T}^{-1}(s_i) \mathbf{\dot{T}}(s_i))^{\vee} = \mathbf{J}_a(s_i,\mathbf{q},\mathbf{w}) \mathbf{\dot{q}} + \sum_{k=1}^p \mathbf{C}^k(s_i,\mathbf{q},\mathbf{w}) \mathbf{\dot{w}}^k
\end{equation}
where $s_i$ is the arc-length location of the $i$-th contact,  $(\cdot)^{\vee}$ denotes the six-dimensional vector form of the $4 \times 4$ matrix twist \citep{lynch2017modern}, and the actuation Jacobian $\mathbf{J}_a$ and the compliance matrix $\mathbf{C}^k$ are defined as 
\begin{subequations}
\begin{align}
    \mathbf{J}_a(s_i,\mathbf{q},\mathbf{w}) & = \left [ \left (\mathbf{T}(s_i)^{-1} \frac{\partial \mathbf{T}(s_i)}{\partial q_1}  \right )^{\vee}, \ldots, \left (\mathbf{T}(s_i)^{-1} \frac{\partial \mathbf{T}(s_i)}{\partial q_{d}} \right )^{\vee} \right ] \label{eq:actuation_jacobian} \\
    \mathbf{C}^k(s_i,\mathbf{q},\mathbf{w}) & = \left [ \left (\mathbf{T}(s_i)^{-1} \frac{\partial \mathbf{T}(s_i)}{\partial w_1^k} \right )^{\vee}, \ldots, \left (\mathbf{T}(s_i)^{-1}  \frac{\partial \mathbf{T}(s_i)}{\partial w_6^k} \right )^{\vee} \right ] \end{align}
\end{subequations}
We remark that these matrices are known when the robot's mechanics-based model is given~\citep{rucker2011_Jacob, 2024CTR_jia}. 

Note that $\mathbf{T}(s_i)$ describes the position and orientation of the robot's center line. Since a contact point $\mathbf{p}^i$ is on the surface of the robot, a local offset from the center line exists, defined as $\mathbf{r}^i \in \Re^3$. The position of the $i$-th contact point is given by
\begin{equation} \label{eq:p_T_r}
    \mathbf{\tilde{p}}^i = \mathbf{T}(s_i) \mathbf{\tilde{r}}^i,
\end{equation}
where $\mathbf{\tilde{p}}^i$ and $\mathbf{\tilde{r}}^i$ are the homogeneous vectors of $\mathbf{p}^i$ and $\mathbf{r}^i$, respectively.
Noting that the contacts exert pure forces without moments, the contact wrench $\mathbf{w}^k$ is the contact force $\mathbf{f}^k$ combined with the moment induced by the offset vector $\mathbf{r}^k$, given by
\begin{equation} \label{eq:def_wk}
    \mathbf{w}^k = \begin{bmatrix}
        \mathbf{r}^k \times \mathbf{f}^k \\ \mathbf{f}^k
    \end{bmatrix} = \begin{bmatrix}
        [\mathbf{r}^k]_{\times} \mathbf{f}^k \\ \mathbf{f}^k
    \end{bmatrix},
\end{equation}
where the contact force $\mathbf{f}^k$ is expressed in the local frame, and $[\cdot]_{\times}$ for a three-dimensional vector denotes the $3 \times 3$ skew-symmetric matrix representation~\citep{lynch2017modern}.

Taking the derivative of (\ref{eq:p_T_r}), (\ref{eq:def_wk}) and plugging into (\ref{eq:T_J_start}), the velocity of the contact point is expressed as
\begin{equation} \label{eq:center_contact}
    \dot{\mathbf{p}}^i = \mathbf{J}_{c,i} \dot{\mathbf{q}} + \sum_{k=1}^p \mathbf{C}_{c,i}^k \dot{\mathbf{f}}^k,
\end{equation}
where $\mathbf{J}_{c,i}$ and $\mathbf{C}_{c,i}^k$ are the contact Jacobian and compliance matrices of the $i$-th contact point, defined as
\begin{equation}
    \mathbf{J}_{c,i} = \mathbf{A}^i \mathbf{J}_a(s_i, \mathbf{q},\mathbf{w}), \ \mathbf{C}_{c,i}^k = \mathbf{A}^i \mathbf{C}^k(s_i, \mathbf{q},\mathbf{w}) \mathbf{B}^k
\end{equation}
with
\begin{equation}
\mathbf{A}^i = \big [ - \mathbf{R}(s_i) [\mathbf{r}^i]_{\times}  ~~ \mathbf{R}(s_i) \big ] , ~ \mathbf{B}^k = \big [ [\mathbf{r}^k]_{\times}^T ~~ \mathbf{I}_3 \big ]^T ,
\end{equation}
where $\mathbf{I}_3$ is the $3 \times 3$ identity matrix.

The displacement of contact points $\mathbf{p}_c$ over a short time interval is given by the Euler integration of (\ref{eq:center_contact}), which is defined as $\Delta \mathbf{p}_c$ by
\begin{equation} \label{eq:pc_J_c}
    \Delta \mathbf{p}_c = \mathbf{J}_c \Delta \mathbf{q} + \mathbf{C}_c \Delta \mathbf{f}_{\text{c}},
\end{equation}
where 
$\mathbf{J}_c$, $\mathbf{C}_c$ denote the combined contact Jacobian and compliance matrices, respectively:
\begin{equation} \label{eq:contact_matrices}
    \mathbf{J}_c = \begin{bmatrix}
        \mathbf{J}_{c,1} \\
        \mathbf{J}_{c,2} \\
        \vdots \\
        \mathbf{J}_{c,p}
    \end{bmatrix}, ~
    \mathbf{C}_c = \begin{bmatrix}
        \mathbf{C}_{c,1}^1 & \mathbf{C}_{c,1}^2 & \cdots & \mathbf{C}_{c,1}^p \\
        \mathbf{C}_{c,2}^1 & \mathbf{C}_{c,2}^2 & \cdots & \mathbf{C}_{c,2}^p \\
        \vdots & \vdots & \ddots & \vdots \\
        \mathbf{C}_{c,p}^1 & \mathbf{C}_{c,p}^2 & \cdots & \mathbf{C}_{c,p}^p
    \end{bmatrix},
\end{equation}
and $\mathbf{f}_{\text{c}}$ is the collection of all contact forces:
\begin{equation} \label{eq:contact_forces}
    \mathbf{f}_{\text{c}} = (\mathbf{f}^1, \ldots, \mathbf{f}^p) \in \Re^{3p}.
\end{equation}

\subsubsection{Contact Mechanics as an LCP}
\label{subsubsec:LCP}
For each contact point, we can consider the corresponding points fixed on the robot and on the environmental surface. In the current state, these two points are aligned with each other.

Suppose the robot actuation and contact forces at the current state are $\mathbf{q}_0$ and  $\mathbf{f}_{0,\text{c}}$, respectively. When the robot is actuated to $\mathbf{q}_0 + \Delta \mathbf{q}$ and the environment has moved to a known position over a small time step, the two points exhibit a relative displacement. These relative displacements of all contact points are collected into a vector $\mathbf{v}_{\text{c}}$, expressed as
\begin{equation} \label{eq:def_v1}
    \mathbf{v}_{\text{c}} = \Delta \mathbf{p}_c - \bm{\rho},
\end{equation}
where $\bm{\rho}$ is the movement of the points in the environment (see Fig.~\ref{fig:delta_p}(b)). By substituting (\ref{eq:pc_J_c}) and defining the updated contact forces as $\mathbf{f}_{\text{c}} \equiv \mathbf{f}_{0, \text{c}} + \Delta \mathbf{f}_{\text{c}}$, Equation (\ref{eq:def_v1}) becomes
\begin{equation} \label{eq:v_h_cw}
    \mathbf{v}_{\text{c}} =  \mathbf{h} + \mathbf{C}_c \mathbf{f}_{\text{c}}, 
\end{equation}
where 
\begin{equation} \label{eq:h}
    \mathbf{h} = -\bm{\rho} + \mathbf{J}_c \Delta \mathbf{q}  -  \mathbf{C}_c \mathbf{f}_{0,\text{c}}. 
\end{equation}
We remark that $\mathbf{h}$ is a known vector based on the actuation, environmental movement, and current contact forces. 

As described in (\ref{eq:f_decompose}), the $k$-th contact force can be decomposed in the normal and tangent directions as 
\begin{equation} \label{eq:w_k_nf_DBeta}
    \mathbf{f}^k = \mathbf{n}^k f_\text{n}^k + 
    \mathbf{D}^k \bm{\beta}^k,
\end{equation}
where $\mathbf{n}^k$ is the normal vector and $\mathbf{D}^k$ is the tangent basis matrix.

By concatenating (\ref{eq:w_k_nf_DBeta}) for all contact points, $\mathbf{f}_{\text{c}}$ is expressed as
\begin{equation} \label{eq:w_nf_Dbeta}
    \mathbf{f}_{\text{c}} = \mathbf{n}_{\text{c}} \mathbf{f}_{\text{n}} + \mathbf{D}_{\text{c}} \bm{\beta}_{\text{c}}
\end{equation}
where $\mathbf{f}_{\text{n}} = (f_\text{n}^1, \ldots, f_\text{n}^p) \in \Re^{p}$, $\bm{\beta}_{\text{c}} = (\bm{\beta}^1, \ldots, \bm{\beta}^p) \in \Re^{mp}$, and $\mathbf{n}_{\text{c}} \in \Re^{3p\times p}$ and $\mathbf{D}_{\text{c}} \in \Re^{3p\times mp}$ are defined as
\begin{equation}
    \mathbf{n}_{\text{c}} = \begin{bmatrix}
        \mathbf{n}^1 & \mathbf{0} & \cdots & \mathbf{0} \\
        \mathbf{0} & \mathbf{n}^2 &  \cdots & \mathbf{0} \\
        \vdots & \vdots & \ddots & \vdots \\
        \mathbf{0} & \mathbf{0} & \cdots & \mathbf{n}^p
    \end{bmatrix}, ~
    \mathbf{D}_{\text{c}} = \begin{bmatrix}
        \mathbf{D}^1 & \mathbf{0} & \cdots & \mathbf{0} \\
        \mathbf{0} & \mathbf{D}^2 &  \cdots & \mathbf{0} \\
        \vdots & \vdots & \ddots & \vdots \\
        \mathbf{0} & \mathbf{0} & \cdots & \mathbf{D}^p
    \end{bmatrix}.
\end{equation}
Plugging (\ref{eq:w_nf_Dbeta}) into (\ref{eq:v_h_cw}), we obtain the relation between $\mathbf{v}_{\text{c}}$ and the contact force components:
\begin{equation} \label{eq:v_fnorm_beta}
    \mathbf{v}_\text{c} = \mathbf{h} + \mathbf{C}_c \mathbf{n}_{\text{c}} \mathbf{f}_{\text{n}} + \mathbf{C}_c \mathbf{D}_{\text{c}} \bm{\beta}_{\text{c}}.
\end{equation}

To formulate the final LCP, we define the variables for CCs (\ref{eq:CCs}) as $\mathbf{v}_n = (v_\text{n}^1, \ldots, v_\text{n}^p) \in \Re^p$, $\bm{\sigma}_\text{c} = (\bm{\sigma}^1, \ldots, \bm{\sigma}^p) \in \Re^{mp}$, $\bm{\eta}_\text{c} = (\eta^1, \ldots, \eta^p) \in \Re^p$, and $\bm{\lambda}_\text{c} = (\lambda^1, \ldots, \lambda^p) \in \Re^p$.
Similar to (\ref{eq:def_vn_fnorm}) and (\ref{eq:CC_tangent}), the following relations hold:
\begin{equation} \label{eq:CCs_mat}
    \begin{aligned}
        \mathbf{v}_{\text{n}} & = \mathbf{n}_{\text{c}}^T \mathbf{v}_{\text{c}} \geq \mathbf{0}, \\
        \bm{\sigma}_{\text{c}} & = \mathbf{e}_{\text{c}} \bm{\lambda}_{\text{c}} + \mathbf{D}_{\text{c}}^T \mathbf{v}_{\text{c}} \geq \mathbf{0}, \\
        \bm{\eta}_{\text{c}} & = \bm{\mu}_{\text{c}} \mathbf{f}_{\text{n}} - \mathbf{e}^T_{\text{c}} \bm{\beta}_{\text{c}} \geq \mathbf{0}
    \end{aligned}
\end{equation}
where 
$\bm{\mu}_{\text{c}} \in \Re^{p \times p}$ and $\mathbf{e}_{\text{c}} \in \Re^{mp\times p}$ are defined as
\begin{equation} \label{eq:def_mu_e}
    \bm{\mu}_{\text{c}} = \begin{bmatrix}
        \mu^1 & 0 & \cdots & 0 \\
        0 & \mu^2 &  \cdots & 0 \\
        \vdots & \vdots & \ddots & \vdots \\
        0 & 0 & \cdots & \mu^p
    \end{bmatrix}, 
    \mathbf{e}_{\text{c}} = \begin{bmatrix}
        \mathbf{e}^1 & \mathbf{0} & \cdots & \mathbf{0} \\
        \mathbf{0} & \mathbf{e}^2 &  \cdots & \mathbf{0} \\
        \vdots & \vdots & \ddots & \vdots \\
        \mathbf{0} & \mathbf{0} & \cdots & \mathbf{e}^p
    \end{bmatrix}
\end{equation}
with $\mathbf{e}^k = (1,\ldots,1) \in \Re^{m}$. Combining (\ref{eq:v_fnorm_beta}) with the CCs in (\ref{eq:CCs_mat}), the LCP with the unknowns $(\mathbf{v}_n, \bm{\sigma}_{\text{c}}, \bm{\eta}_{\text{c}})$ and $ (\mathbf{f}_{\text{n}}, \bm{\beta}_{\text{c}}, \bm{\lambda}_{\text{c}})$ is formulated as:
\begin{subequations} \label{eq:LCP_general}
    \begin{align}
        \text{LCP}(\mathbf{M}, \mathbf{\tilde{h}}) & : ~ 
        \begin{bmatrix}
            \mathbf{v}_\text{n} \\ \bm{\sigma}_{\text{c}} \\ \bm{\eta}_{\text{c}}
        \end{bmatrix} =
        \mathbf{M}
        \begin{bmatrix}
            \mathbf{f}_{\text{n}} \\ \bm{\beta}_{\text{c}} \\ \bm{\lambda}_{\text{c}}
        \end{bmatrix} + 
        \mathbf{\tilde{h}} \label{eq:LCP_general_relation} \\
         & \text{s.t.} ~ ~ \mathbf{0} \leq \begin{bmatrix}
            \mathbf{v}_\text{n} \\ \bm{\sigma}_{\text{c}} \\ \bm{\eta}_{\text{c}}
        \end{bmatrix} \perp \begin{bmatrix}
            \mathbf{f}_{\text{n}} \\ \bm{\beta}_{\text{c}} \\ \bm{\lambda}_{\text{c}}
        \end{bmatrix} \geq \mathbf{0}, \label{eq:LCP_general_cc}
    \end{align}
\end{subequations}
where 
\begin{equation} \label{eq:def_M_h}
    \mathbf{M} = \begin{bmatrix}
            \mathbf{n}^T_{\text{c}} \mathbf{C}_c \mathbf{n}_{\text{c}} & \mathbf{n}_{\text{c}}^T \mathbf{C}_c \mathbf{D}_{\text{c}} & \mathbf{0} \\
            \mathbf{D}^T_{\text{c}} \mathbf{C}_c \mathbf{n}_{\text{c}} & \mathbf{D}^T_{\text{c}} \mathbf{C}_c \mathbf{D}_{\text{c}} & \mathbf{e}_{\text{c}} \\
            \bm{\mu}_{\text{c}} & -\mathbf{e}^T_{\text{c}} & \mathbf{0}
        \end{bmatrix}, ~
    \mathbf{\tilde{h}} = 
        \begin{bmatrix}
            \mathbf{n}^T_{\text{c}} \mathbf{h} \\ \mathbf{D}^T_{\text{c}} \mathbf{h} \\  \mathbf{0}
        \end{bmatrix}.
\end{equation}

Here, we prove the existence of the solution for the LCP in (\ref{eq:LCP_general}). Supposing that $\mathbf{z} = [\mathbf{f}_{\text{n}}, \bm{\beta}_{\text{c}}, \bm{\lambda}_{\text{c}}] \geq \mathbf{0}$, we can verify that $\mathbf{M}$ is copositive by: 
\begin{equation} \label{eq:zMz0}
    \mathbf{z}^T \mathbf{M} \mathbf{z} = \frac{1}{2} \mathbf{z}^T (\mathbf{M}+\mathbf{M}^T) \mathbf{z} =
    \mathbf{f}_{\text{c}}^T \mathbf{C}_c \mathbf{f}_{\text{c}} + \mathbf{f}_{\text{n}}^T \bm{\mu}_{\text{c}} \bm{\lambda}_{\text{c}}
     \geq 0,
\end{equation}
where the final inequality holds because $\bm{\mu}_{\text{c}} > 0$ and the compliance matrix $\mathbf{C}_c$ is positive definite in stable configurations. 

Subsequently, we suppose that $\mathbf{z} \in \text{LCP}(\mathbf{M}, \mathbf{0})$, thus $\mathbf{z}^T \mathbf{M} \mathbf{z} = 0$ due to the complementarity constraints. From (\ref{eq:zMz0}), we have $\mathbf{f}_{\text{c}} = \mathbf{0}$, which further implies that
\begin{equation}
    \mathbf{\tilde{h}}^T \mathbf{z} = \mathbf{h}^T \mathbf{n}_{\text{c}} \mathbf{f}_n + \mathbf{h}^T \mathbf{D}_{\text{c}} \bm{\beta}_{\text{c}} = \mathbf{h}^T \mathbf{f}_{\text{c}} = \mathbf{0}
\end{equation}
Following Lemma 1, the solution set of the LCP in (\ref{eq:LCP_general}) is non-empty.

After solving the LCP, the contact forces $\mathbf{f}_{\text{c}}$ and the contact location changes $\Delta \mathbf{p}_c$ are determined by (\ref{eq:w_nf_Dbeta}) and (\ref{eq:pc_J_c}), respectively. The robot shape can be reconstructed by the mechanics-based model in (\ref{eq:ode_cosserat}).

\subsubsection{Compensation for Penetration and Disturbance}

The LCP formulation in (\ref{eq:LCP_general}) was derived for ideal situations where the robot is in exact contact with the environmental surface. In practice, contact depth errors often arise due to the numerical errors caused by linearizations (e.g., penetration induced by planar approximation of a curved contact surface). These errors can be corrected by modifying the first inequality of (\ref{eq:CCs_mat}) to $\mathbf{n}_{\text{c}}^T \mathbf{v}_{\text{c}} \geq \bm{\epsilon}$, where $\bm{\epsilon}$ collects the contact depth errors. The resulting constraint must be shifted into standard LCP form. Equivalently, a depth correction vector $\mathbf{b}$ can be added to $\mathbf{h}$ in (\ref{eq:v_h_cw}) as 
\begin{equation}
    \mathbf{h} \leftarrow \mathbf{h} + \mathbf{b},
    \label{eq:b}
\end{equation} 
where $\mathbf{b}$ is illustrated in Fig.~\ref{fig:contact_type}. 
In addition to the contact forces considered in (\ref{eq:pc_J_c}), the robot's deformation due to other non-contact external loads, such as environmental disturbances or active actuation forces, can also be accounted for by adding another term in (\ref{eq:v_h_cw}) as $\mathbf{h} \leftarrow \mathbf{h} + \mathbf{C}_{\text{ext}} \Delta \mathbf{f}^{\text{ext}}$, where $\mathbf{f}^{\text{ext}}$ and $\mathbf{C}_{\text{ext}}$ denote the external loads and the associated compliance matrices.

\subsection{Contact Detection} \label{sec:contact_detect}
The proposed LCP formulation requires the knowledge of contact locations $\mathbf{p}_c$ and normal directions $\mathbf{n}$. We assume that the geometry of the environment and the robot's shape $\mathbf{p}(s)$ are known at $t=t_j$. The problem of contact detection is then defined as:
\begin{equation} \label{eq:prob_contact_detect}
\begin{aligned}
    \text{Find} & ~~ (s, \mathbf{r}), \\
    \text{subject to} & ~~ \Phi(\mathbf{p}(s)+\mathbf{R}(s) \mathbf{r})=0,
\end{aligned}
\end{equation}
where $\mathbf{r}$ is a vector of the local offset from the center line as defined in (\ref{eq:p_T_r}), and 
$\Phi(\cdot)$ denotes the signed distance field (SDF) of a point on the robot's surface to the environment. The location of a detected contact at $s_i$ is $\mathbf{p}^i=\mathbf{p}(s_i)+\mathbf{R}(s_i)\mathbf{r}^i$. In this work, we focus on cases where $\Phi(\cdot)$ admits an analytic form (e.g., planar or cylindrical geometries). For arbitrary environments, $\Phi(\cdot)$ can be approximated using numerical signed distance fields constructed from sensor data or mesh-based representations.
An operator $\Lambda(\cdot)$ is used to represent the solution of (\ref{eq:prob_contact_detect}) given the robot's configuration $\mathbf{T}(s)$ and the location of the environment: $\mathcal{C} = \Lambda(\mathbf{T}(s), t)$, where $\mathcal{C}$ is the set of detected pairs $(s,\mathbf{r})$. The time variable $t \in \Re$ is introduced as the state of the moving environment depends on $t$.
Subsequently, the number of contacts ($p$) is given as the cardinality of $\mathcal{C}$.

\begin{figure} 
    \centering
    \includegraphics[width = 0.9\linewidth]{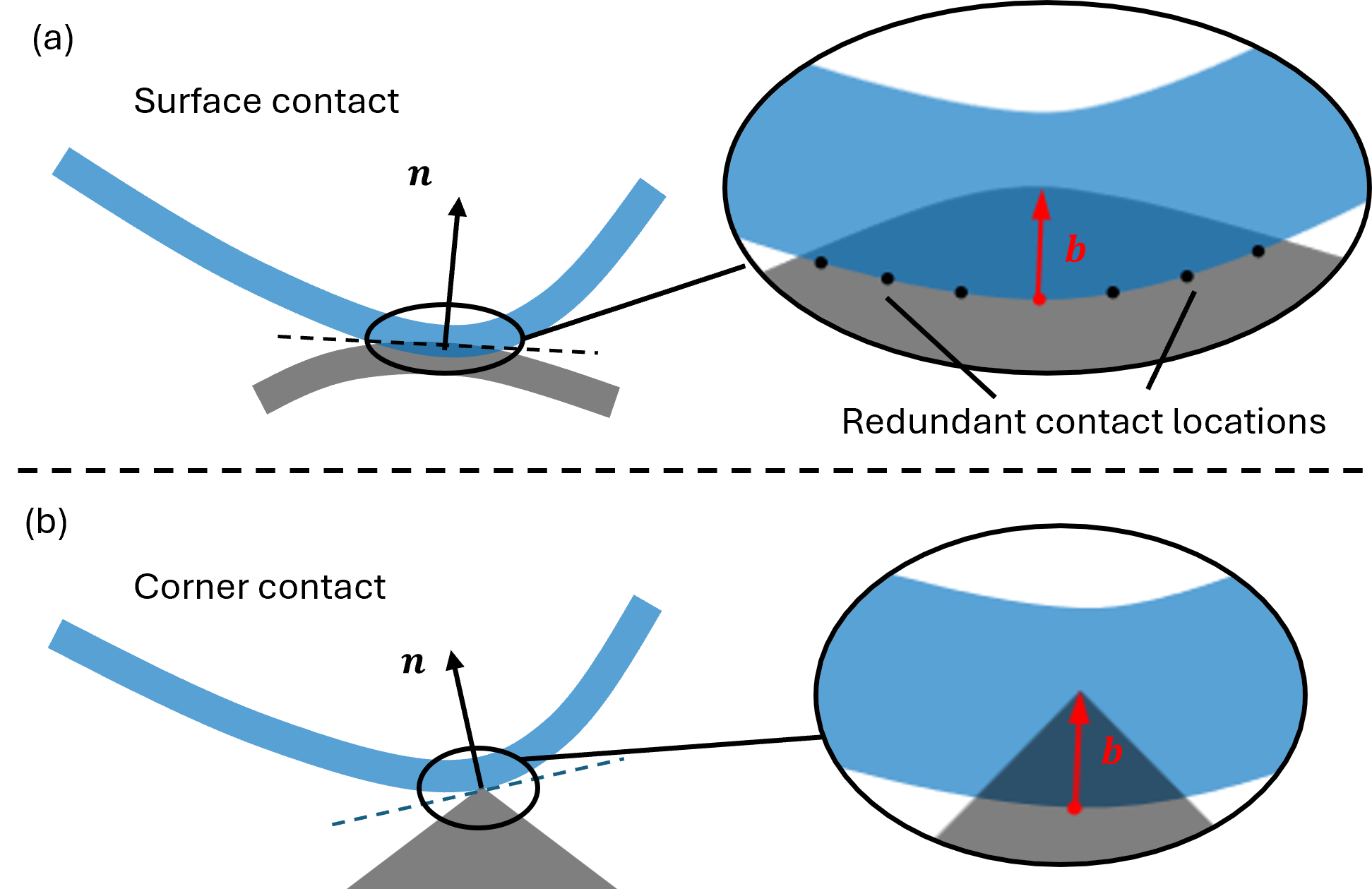}
    \caption{Geometric representation for two types of contact. (a) Surface contact: $\mathbf{n}$ depends on the surface. (b) Corner contact: $\mathbf{n}$ depends on the shape of the rod. In each type of contact, the depth correction vector $\mathbf{b}$ with the largest penetration is shown in zoom-in plots.
    }
    \label{fig:contact_type}
\end{figure}

In our implementation, contact detection is performed over discretized points along the robot's arc length. Note that the robot in our computation may experience small penetrations due to numerical issues, which leads to $\Phi<0$ and consecutive contact locations for a single contact as seen in Fig.~\ref{fig:contact_type}(a). Since the Jacobian and compliance matrices at these consecutive arc-length locations are close to each other, the resulting LCP formulation (\ref{eq:LCP_general}) becomes near-singular and thus ill-conditioned. 
To address this numerical issue, we assume that all contacts are rigid and select a single representative location $s_i$ with the largest penetration, discarding adjacent points (e.g., $s_{i-1}$, $s_{i+1}$, ...).


\corrlab{R1-2}{
With the contact locations detected, the normal $\mathbf{n}^i$ at contact $i$ is assigned according to two contact patterns, as illustrated in Fig.~\ref{fig:contact_type}. For surface contact, where the robot interacts with a smooth environmental surface, the normal direction is obtained from the normalized signed distance field (SDF) gradient at the contact point,
\begin{equation}
\mathbf{n}^i
=
\frac{\nabla_{\mathbf{p}} \Phi(\mathbf{p}^i)}
{\left|| \nabla_{\mathbf{p}} \Phi(\mathbf{p}^i) \right||},
\end{equation}
assuming that $\Phi$ is defined such that its gradient points outward from the obstacle. For corner contact, where the robot body interacts with a sharp feature such as the edge or vertex of a polygonal obstacle, the SDF gradient may not be uniquely defined. A fully geometric treatment would require resolving the local microscopic structure of the sharp feature or introducing an explicit smoothing model, such as a small-radius fillet. In this work, we instead exploit the fact that the robot body has a smooth circular cross-section. The contact normal is therefore taken as the inward normal of the robot surface at the contact point, expressed in the global frame: $\mathbf{n}^i = -\mathbf{R}(s_i)\mathbf{r}^i/{||\mathbf{r}^i||}.
$ This choice provides a well-defined normal direction even when the environmental normal is ambiguous.
}

\subsection{Algorithm}  \label{sec:algorithm}

Lastly, we outline our algorithm for determining the robot shape and the contact forces/locations over discretized time steps. We assume that the robot's configuration at the current time step $t=t_j$, $\mathbf{T}_j(s)$, is known, along with the contact forces and locations. The algorithm consists of a loop that updates these quantities for the subsequent time step $t=t_{j+1}$ through the following two phases: (i) solving the LCP and (ii) detecting and correcting contacts.

In the first phase, the LCP (\ref{eq:LCP_general}) is solved to update the contact locations and forces. To acquire the variables necessary for the LCP, the Jacobian $\mathbf{J}_c$ and the compliance matrix $\mathbf{C}_c$ at $t = t_j$ are computed using the mechanics-based model as described in \hyperref[subsubsec:diff_kin]{Differential kinematics} section. The vector $\mathbf{h}$ is then computed by (\ref{eq:h}) based on the actuation and environmental changes, as well as the current contact forces. This vector is further corrected by (\ref{eq:b}) when contact depth errors exist. 
For surface contact (Figure~\ref{fig:contact_type}), the variables $\mathbf{n}$ and $\mathbf{D}$ are obtained from the environmental geometry at the contact points. For corner contact, where the environment normal is not uniquely defined, these variables are computed from the robot surface normal at the previous time step and kept fixed within each iteration.
Once the LCP is solved, the robot's configuration is updated to $\mathbf{T}_{j+1}(s)$.

The solution of the LCP determines whether a certain contact is maintained or broken. Contacts with zero force ($\mathbf{f}^k=\mathbf{0}$) and non-zero normal take-off ($v_n^k > 0$) are eliminated. For the remaining active contacts, the LCP solution also provides the new contact locations and forces. These locations may exhibit numerical drifts due to the linearization assumptions. Furthermore, new contacts can be introduced by the small shape change of the robot.

For these reasons, in the second phase, we detect all the contacts again by $ \mathcal{C}_{j+1} = \Lambda(\mathbf{T}_{j+1}(s), t_{j+1})$.
For the contacts in $\mathcal{C}_{j+1}$ that correspond to the existing active contacts, we use the newly detected locations. The other contacts in $\mathcal{C}_{j+1}$ are the new contacts, to which we assign zero contact forces. The combination of these forces constitutes the new contact force vector $\mathbf{f}_{j+1}$.

\begin{figure} 
    \centering
    \includegraphics[width = \linewidth]{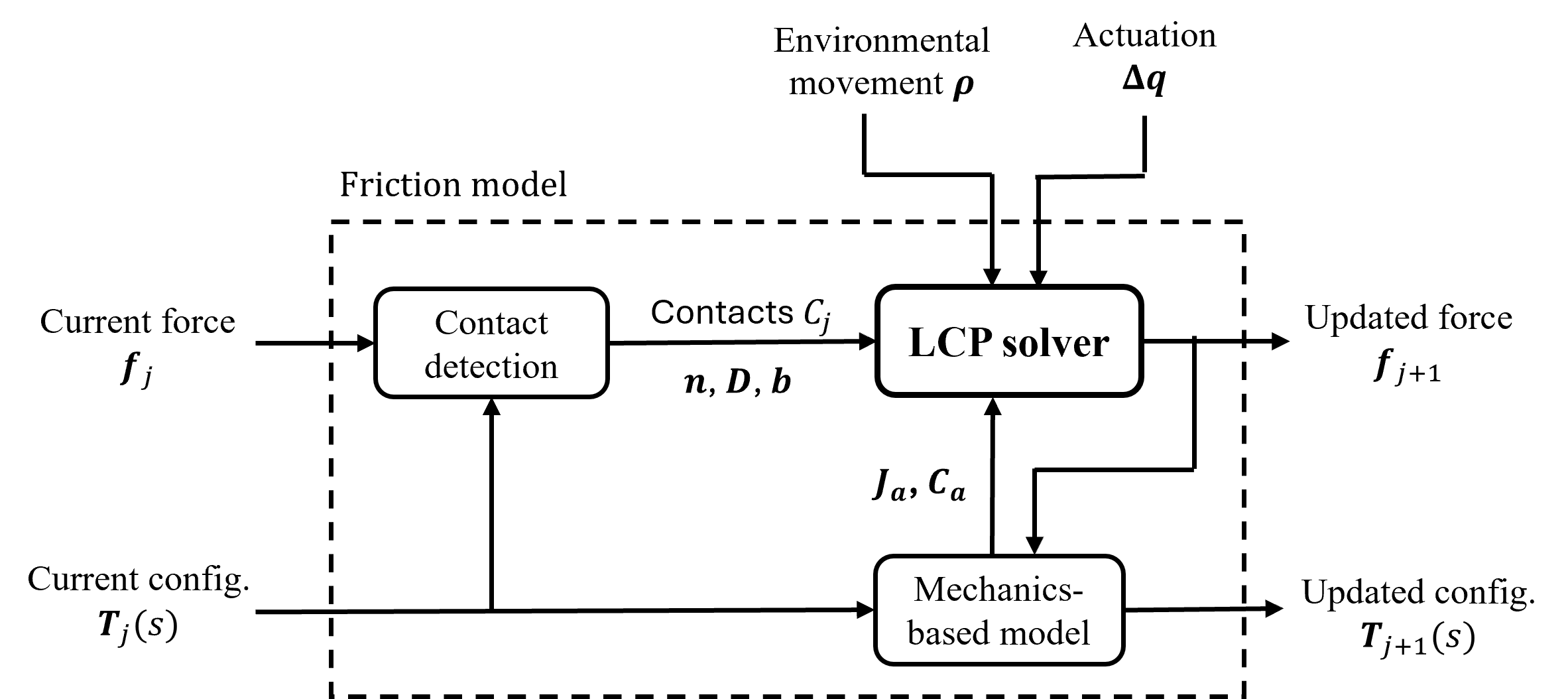}
    \caption{Flow chart of steps for detecting the contacts and solving friction. For the initial configuration $\mathbf{T}_0(s)$, we preferably use a contactless configuration.
    }
    \label{fig:flowchart}
\end{figure}

The flow chart in Figure~\ref{fig:flowchart} and Algorithm \ref{alg:contact_detect} summarize the steps for friction modeling and contact detection. In Algorithm~\ref{alg:contact_detect}, the solution of the LCP formulation in (\ref{eq:LCP_general}), given the contact information $(\mathcal{C}, \mathbf{f})$, is denoted as LCP$(\mathcal{C}, \mathbf{f})$. The zero vector in line $10$ corresponds to the contact forces of the new contacts. Since the initial contact information $(\mathcal{C}_0, \mathbf{f}_0)$ is generally unavailable, we preferably execute our algorithm with a contact-free initial configuration in our later simulations. In this case, no LCPs need to be solved and the robot's configuration is updated through frictionless kinematic models until contacts are detected. 

To ensure the robustness of our algorithm, the assumptions outlined in \hyperref[sec:lcp_general]{LCP Formulation of external friction} section must be strictly upheld. Excessively large step lengths can invalidate the use of local Jacobians and compliance matrices in (\ref{eq:pc_J_c}), while the local planar approximation in assumption (i) may be undermined when the robot interacts with highly curved surfaces. In either case, contact points from the previous step may be incorrectly removed, potentially resulting in oscillations or other unrealistic behaviors.

\begin{algorithm}
\caption{Contact detection and friction modeling}\label{alg:contact_detect}
\begin{algorithmic}[1]
\State \textbf{Initialize} $\mathbf{T}_0(s)$ and $(\mathcal{C}_0, \mathbf{f}_0)$
\State $j \gets 0$
\While{$j \leq N
$}
\If{$\mathcal{C}_j = \{ \}$} 
    \State $\mathbf{T}_{j+1}(s) \gets \mathcal{F}(\mathbf{q}_{j+1}, \mathbf{0})$.
\Else
    \State Solve LCP (\ref{eq:LCP_general}): $(\mathcal{C}_{j+1}, \mathbf{f}_{j+1}) \gets \text{LCP}(\mathcal{C}_{j}, \mathbf{f}_{j})$.
    \State $\mathbf{T}_{j+1}(s) \gets \mathcal{F}(\mathbf{q}_{j+1}, \mathbf{f}_{j+1})$.
\EndIf 
    \State Contact detection: $\mathcal{C}_{j+1} \gets \Lambda(\mathbf{T}_{j+1}(s), t_{j+1})$.
    \State Contact force update: $\mathbf{f}_{j+1} \gets (\mathbf{f}_{j+1}, \mathbf{0})$.
\State $j \gets j+1$.
\EndWhile
\end{algorithmic}
\end{algorithm}

\section{Case Study: Application to an Elastic Rod}
\label{sec:rod}
In this section, we instantiate the general formulation for a simple continuum robot, namely an elastic rod actuated at its base. 
We derive the explicit forms of the contact-related matrices within the Cosserat rod framework, enabling direct implementation of the proposed model.
We further revisit the frictionless contact kinematics from our previous work~\citet{Previous_work} and establish its equivalence to the proposed formulation in the absence of friction.

\subsection{Rod Kinematics with Frictional Contacts}
\label{sec:review_rod}

The kinematics of an elastic rod can be formulated as a finite-dimensional problem by discretizing the center-line position and curvature along the arc-length and concatenating them into vectors:
\begin{equation}
\label{eq:u_p_discret}
    \mathbf{\underline{u}} = [\mathbf{u}(s_1)^T, \ldots, \mathbf{u}(s_{n})^T]^T ,~~~ 
    \mathbf{\underline{p}} = [\mathbf{p}(s_1)^T, \ldots, \mathbf{p}(s_{n})^T]^T,
\end{equation}
where $s_i$ is the arc-length location of the $i$-th discretization point along the rod. A similar concatenation notation is used for the precurvature and the stiffness matrix along the rod:
\begin{equation}
    \textbf{\underline{K}} = \text{diag}(\mathbf{K},\ldots,\mathbf{K}) , ~~~
    \mathbf{\underline{\hat{u}}} = [\mathbf{\hat{u}}(s_1)^T, \ldots, \mathbf{\hat{u}}(s_n)^T]^T.
\end{equation}

Here, we applied the proposed friction model to the external friction of an elastic rod, which involves deriving $\mathbf{h}$ and $\mathbf{C}_c$ in (\ref{eq:v_h_cw}).
We consider a single contact at $s_i$ for simplicity in derivation, i.e., $\mathbf{p}_c = \mathbf{p}(s_i)$. Multiple contacts can be incorporated straightforwardly using matrices and force vectors defined in (\ref{eq:contact_matrices})--(\ref{eq:contact_forces}).
Additionally, the radius of the rod is assumed to be small such that $\mathbf{r} = \mathbf{0}$ in (\ref{eq:p_T_r}). 

First, the compliance matrix $\mathbf{C}_c$ can be derived by considering zero actuation in (\ref{eq:pc_J_c}), leading to $\Delta \mathbf{p}(s_i) = \mathbf{C}_c \Delta \mathbf{f}$. 
The position Jacobian of the contact point, $\mathbf{J}_\mathbf{u} \in \Re^{3\times 3n}$, can be used to compute the small position displacement for a small curvature change:
\begin{equation} \label{eq:p_J_u}
    \Delta \mathbf{p}(s_i) = \mathbf{J}_\mathbf{u} \Delta \mathbf{\underline{u}}.
\end{equation}
We remark that $\mathbf{J}_\mathbf{u}$ relates the shape change to the mechanics of the rod, different from the actuation Jacobian $\mathbf{J}_a$ defined in (\ref{eq:actuation_jacobian}). The analytical form of $\mathbf{J}_\mathbf{u}$ can be obtained by taking the derivative of (\ref{eq:p_u_relation}), which is detailed in \citet{ctr_friction_ha2019}. Here we use a linear approximation of $\mathbf{J}_\mathbf{u} = [\textbf{J}_{\mathbf{u},1}, \ldots, \textbf{J}_{\mathbf{u},n}]$ with the $j$-th block $\mathbf{J}_{\mathbf{u},j} \in \Re^{3\times 3}$, defined in \citet{ha2017incorporating} as:
\begin{equation}
\label{eq:analytical_J}  
    \mathbf{J}_{\mathbf{u},j} =\left\{ \begin{aligned}&[\mathbf{p}(s_{j})-\mathbf{p}(s_{i})]_{\times}\mathbf{R}(s_{j}) (s_{j+1} - s_{j}) , ~ \text{~if~} j \leq i \\
    &\textbf{0}, ~ \text{~if~} j > i
    \end{aligned}\right. .
\end{equation}

Note that the contact force projected through $\mathbf{J}_\mathbf{u}^T$ yields the cross-sectional moments of the rod at mechanical equilibrium~\citep{ctr_friction_ha2019}. We consider a contact force change from $\mathbf{f}_0$ to $\mathbf{f}$, causing the corresponding curvature change from $\mathbf{\underline{u}}_0$ to $\mathbf{\underline{u}}_0 + \Delta \mathbf{\underline{u}}$. Employing the linear constitutive law, the cross-sectional moments equilibrium before and after the force change are given by
\begin{subequations} \label{eq:force_balance}
\begin{align}
    & \mathbf{\underline{K}} (\mathbf{\underline{u}}_0 -  \mathbf{\underline{\hat{{u}}}})  = \mathbf{J}_\mathbf{u}^T \mathbf{f}_0, \label{eq:force_balance_a} \\
    & \mathbf{\underline{K}} (\mathbf{\underline{u}}_0 + \Delta \mathbf{\underline{u}} - \mathbf{\underline{\hat{{u}}}}) = \mathbf{J}_\mathbf{u}^T \mathbf{f}. \label{eq:force_balance_b}
\end{align}
\end{subequations}
Subtracting (\ref{eq:force_balance_a}) from (\ref{eq:force_balance_b}) results in $\Delta \mathbf{\underline{u}}$ as
\begin{equation} \label{eq:delta_u}
    \Delta \mathbf{\underline{u}} = \mathbf{\underline{K}}^{-1} \mathbf{J}_\mathbf{u}^T ( \mathbf{f} - \mathbf{f}_0).
\end{equation}
Finally, a linear relation between $\Delta \mathbf{p}(s_i)$ and $\mathbf{f}$ is obtained by combining (\ref{eq:p_J_u}) and (\ref{eq:delta_u}):
\begin{equation} \label{eq:relation_v_f}
    \Delta \mathbf{p}(s_i) = \mathbf{J}_\mathbf{u} \mathbf{\underline{K}}^{-1} \mathbf{J}_\mathbf{u}^T (\mathbf{f} - \mathbf{f}_0).
\end{equation}
Comparing (\ref{eq:relation_v_f}) with $\Delta \mathbf{p}(s_i) = \mathbf{C}_c \Delta \mathbf{f}$, the compliance matrix is given by 
\begin{equation} \label{eq:C_tube}
    \mathbf{C}_c = \mathbf{J}_\mathbf{u} \mathbf{\underline{K}}^{-1} \mathbf{J}_\mathbf{u}^T.
\end{equation}
The vector $\mathbf{h}$ in (\ref{eq:h}) is also known once $\mathbf{J}_c \Delta \mathbf{q}$ is determined, which corresponds to the movement of $\mathbf{p}(s_i)$ according to the robot's actuation. As our actuation is a base transformation, rather than deriving $\mathbf{J}_c$ separately, we compute $\mathbf{J}_c \Delta \mathbf{q}$ directly as the displacement of $\mathbf{p}(s_i)$ induced purely by the base transformation.

The LCP of an elastic rod is then formulated by plugging the derived $\mathbf{C}_c$ and $\mathbf{h}$ into the general LCP (\ref{eq:LCP_general}). Afterward, the contact force $\mathbf{f}$ and the curvature change $\Delta \mathbf{\underline{u}}$ are computed using 
(\ref{eq:w_nf_Dbeta}) and (\ref{eq:delta_u}), respectively. The shape of the rod is then reconstructed by integrating (\ref{eq:p_u_relation}) using the changed curvature $\mathbf{\underline{u}}_0 + \Delta \mathbf{\underline{u}}$. 

\subsection{Equivalence to Frictionless Contact Model} 
\label{sec:math_eq}

The proposed model remains applicable even in the absence of friction at the contacts. In this case, the model should align with previously established frictionless models. To demonstrate this, we provide a mathematical proof showing the equivalence between the proposed friction model and the linearized energy-based kinematic model under the assumption of no friction.


Once again, we consider a single contact at $s_i$ for simplicity. Following our previous work in \citet{Previous_work}, the shape of the rod under static equilibrium is obtained by minimizing the potential energy subject to the contact constraints: 
\begin{subequations} \label{eq:optim_prob_incre}
    \begin{align}
        \underset{\Delta  \mathbf{\underline{u}}}{\min} ~~ & E_p(\mathbf{\underline{u}}_0+\Delta \mathbf{\underline{u}}, \mathbf{\hat{\underline{u}}}, \mathbf{\underline{K}}) 
        \label{eq:optim_obj_incre}\\
         \text{s.t.} ~~ & - \Phi(\mathbf{p}(s_i)) \leq 0, \label{eq:optim_constraint_incre} \\
        & \mathbf{p}(s_i) = F_i(\mathbf{\underline{u}}_0+\Delta \mathbf{\underline{u}}, \mathbf{T}(0)), \label{p_u_F}
    \end{align}
\end{subequations}
where $\mathbf{\underline{u}}_0$ is the curvature at the last time step, $\Delta \mathbf{\underline{u}}$ is the change of curvature during the current time step, and  $\Phi(\mathbf{p}(s_i))$ is the distance from the $i$-th discretized point to the environment. $F(\cdot)$ integrates (\ref{eq:p_u_relation}) to calculate $\mathbf{\underline{p}}$ using $\mathbf{\underline{u}}$ with a base transformation $\mathbf{T}(0)$, and the potential energy stored in the rod is expressed as:
\begin{equation} \label{eq:E_p}
   E_p(\mathbf{\underline{u}}, \mathbf{\underline{\hat{u}}}, \textbf{\underline{K}}) = \frac{1}{2} (\mathbf{\underline{u}}-\mathbf{\underline{\hat{u}}})^{T}\textbf{\underline{K}}(\mathbf{\underline{u}}-\mathbf{\underline{\hat{u}}}).
\end{equation}

Let $\Delta \mathbf{\underline{u}}^{\star}$ denote the primal solution and $\bm{\xi}^{\star}$ denote the dual solution of (\ref{eq:optim_prob_incre}), with $\xi_i=(\bm{\xi}^{\star})_i$ denoting the Lagrange multiplier of the inequality~(\ref{eq:optim_constraint_incre}) at $s_i$. The Karush–Kuhn–Tucker (KKT) conditions~\citep{boyd2004convex} are then enforced as
\begin{subequations} \label{eq:kkt}
    \begin{align}
        & \nabla_{(\Delta \mathbf{\underline{u}})} E_p(\mathbf{\underline{u}}_0+\Delta \mathbf{\underline{u}}^{\star}, \mathbf{\hat{\underline{u}}}, \mathbf{\underline{K}}) 
        - \xi_i \cdot \nabla_{(\Delta \mathbf{\underline{u}})} \Phi(\mathbf{p}(s_i)) = 0 \label{eq:kkt1} \\ 
        & \xi_i \cdot \Phi(\mathbf{p}(s_i)) = 0, \label{eq:kkt2} \\
        & - \Phi(\mathbf{p}(s_i)) \leq 0, \label{eq:kkt3} \\
        & \xi_i \geq 0. \label{eq:kkt4}
    \end{align}
\end{subequations}

The contact distance function is given by $\Phi(\mathbf{p}(s_i)) = \mathbf{n}^T \mathbf{v}$, where $\mathbf{v}$ is defined in (\ref{eq:v_h_cw}). By substituting (\ref{eq:force_balance_b}) and (\ref{eq:C_tube}), we can equivalently express $\mathbf{v}$ as
$\mathbf{v} = \mathbf{h} + \mathbf{J}_\mathbf{u}(\mathbf{\underline{u}}_0 + \Delta \mathbf{\underline{u}} - \mathbf{\underline{\hat{u}}})$, and the contact distance function becomes
\begin{equation} \label{eq:def_phi}
    \Phi(\mathbf{p}(s_i)) = \mathbf{n}^T \mathbf{h} + \mathbf{n}^T \mathbf{J}_\mathbf{u} (\mathbf{\underline{u}}_0 - \mathbf{\underline{\hat{u}}}) + \mathbf{n}^T \mathbf{J}_\mathbf{u} \Delta \mathbf{\underline{u}}. 
\end{equation}
Taking the derivative of (\ref{eq:optim_obj_incre}), (\ref{eq:def_phi}) w.r.t. $\Delta \mathbf{\underline{u}}$ and plugging into (\ref{eq:kkt1}), we obtain:
\begin{equation} \label{eq:kkt1_deriv}
    \mathbf{\underline{K}} (\mathbf{\underline{u}}_0 + \Delta \mathbf{\underline{u}}^{\star} - \mathbf{\underline{\hat{{u}}}}) = \mathbf{J}_\mathbf{u}^T (\xi_i \mathbf{n}).
\end{equation}
Comparing the above equation with (\ref{eq:force_balance_b}) implies $\xi_i = f_\text{n}$. The physical meaning is that, once the solution to (\ref{eq:optim_prob_incre}) is found, the Lagrange multipliers of the constraint (\ref{eq:optim_constraint_incre}) correspond to the normal contact forces. We remark that a similar conclusion has been drawn in \citet{ctr_friction_ha2019}, despite some different definitions of the constraints.

In frictionless scenarios, the CCs in the tangent directions, (\ref{eq:CCs_friction1}) and (\ref{eq:CCs_friction2}), can be neglected. The CCs of the LCP (\ref{eq:LCP_general_cc}) then reduce to (\ref{eq:CC_origin}), which is equivalent to the KKT conditions in (\ref{eq:kkt2})-(\ref{eq:kkt4}), given that $\xi_i = f_\text{n}$ and $\Phi(\mathbf{p}(s_i)) = \mathbf{n}^T \mathbf{v} = v_\text{n}$.

\section{Numerical Simulations} \label{sec:simu}

In this section, we evaluate the proposed model through simulations of an elastic rod interacting with rigid planes and pipes, followed by a CTR interacting with a plane. The rod is 150 mm long, with a 50 mm precurved distal segment. Its geometric and mechanical parameters are summarized in Table~\ref{table_simu_param}, with Poisson’s ratio $\nu$ determining the ratio of bending to torsional stiffness, $1+\nu$. We also assess the computational efficiency of the model. Animations of the simulations are provided in \hyperref[appendix]{Extension 1}.

\begin{table}[h!]
\renewcommand{\arraystretch}{1.25}
\caption{Rod Parameters in the Simulations}
\label{table_simu_param}
\small\sf\centering
\begin{tabular}{c c c c c }
\toprule
Diameter & Total Len. & Curved Len. & Precurvature & $\nu$ \\
$[\text{mm}]$ & $[\text{mm}]$ & $[\text{mm}]$ & $[\text{mm}^{-1}]$ & --\\
\midrule
1.32 & 150 & 50 & 0.01 & 0.3 \\
\bottomrule
\end{tabular}
\vspace{- 2mm}
\end{table}

\subsection{Simulation 1: Rod-plane Contact} 
\label{sec:simu_case1}

\begin{figure}[t!]
    \centering
    \includegraphics[width = \linewidth]{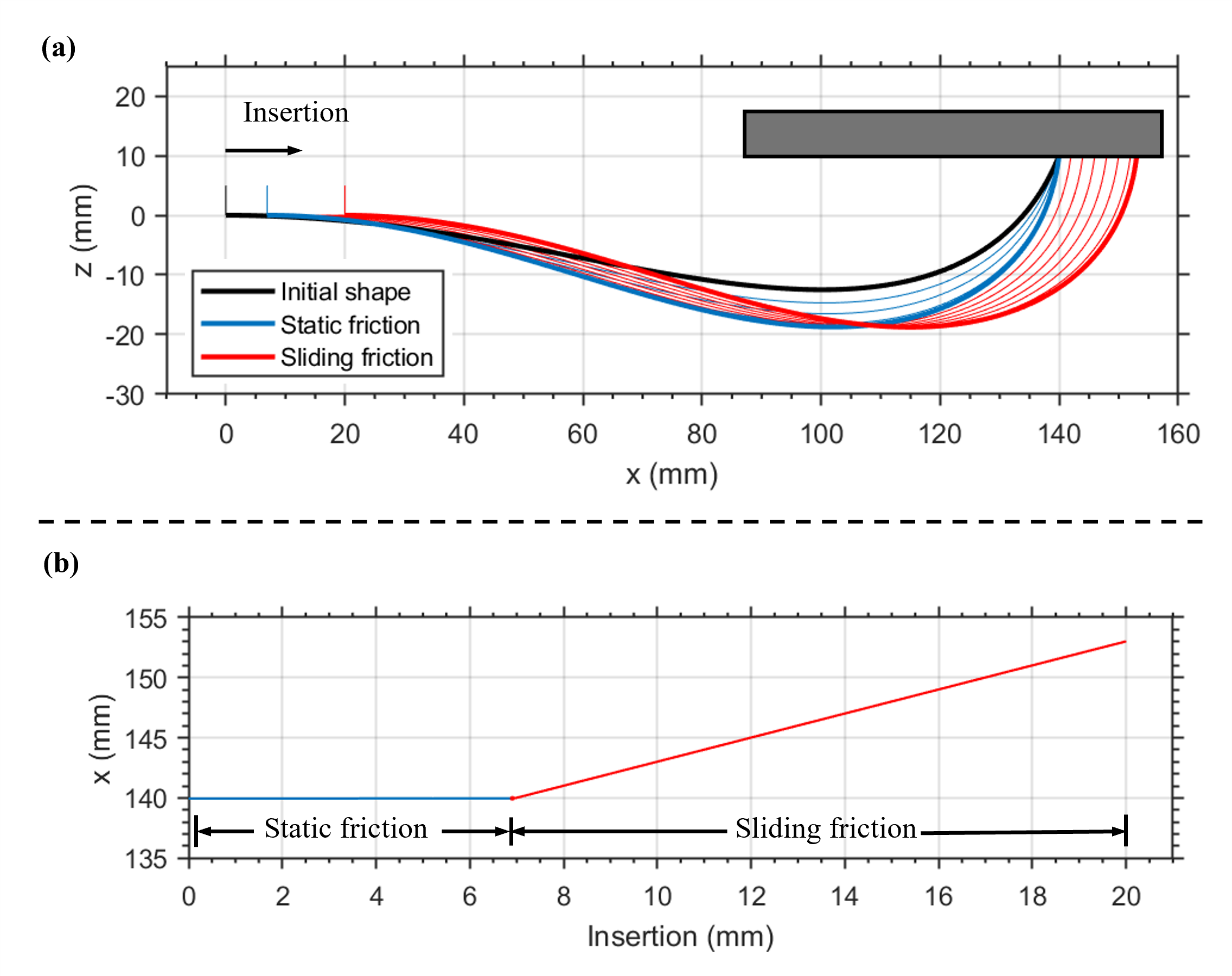}
    \caption{Simulation of a rod in contact with a frictional plane. The rod was inserted forward until relative sliding occurred. (a) The shape trajectory of the rod. (b) The change of the $x$-axis component of the tip position as the rod is inserted. 
    }
    \label{fig:simu_case1}
\end{figure}

To demonstrate the friction model, we first simulated a rod with its tip sliding on a frictional plane. The rod base was inserted along its longitudinal direction, and the tip maintained contact with the plane. The shape trajectory of the rod is shown in Fig. \ref{fig:simu_case1}(a). The tip of the rod first remained stationary over the initial insertion, where the body was being squeezed due to the increasing static friction. As the base was further inserted, the relative sliding between the rod and the plane occurred, and the rod started translating with a constant shape. We term these different contact behaviors as ``contact phases'', with ``static friction phase'' representing the stick motion at the contact ($\mathbf{v} = \mathbf{0}$), and ``sliding friction phase'' indicating the behavior of the rod when relative sliding occurs ($\mathbf{v} \neq \mathbf{0}$). The shift between the contact phases during insertion is shown in Fig. \ref{fig:simu_case1}(b). Note that a large COF of $2.8$ was used in this simulation to manifest stick motion at the tip. 

\subsection{Simulation 2: Rod-pipe Contact} 
\label{sec:simu_case2} 
Next, we simulated the interactions between a precurved inner rod and a straight, rigid outer pipe. This rod-pipe design is considered a specialized case of a CTR and is commonly encountered in practical applications, such as treatment of intracerebral hemorrhage (ICH) \citep{jessica2013debulk, Anthony2024ICH} and continuum robot-based pipe inspection \citep{Previous_work, yu2022design}. While our model is not directly applicable to the internal tube-tube friction of general CTRs, studying the rod-pipe scenario provides valuable insights into the frictional contacts between the elastic rod and long, thin cylindrical surfaces.

\begin{figure}[t]
    \centering
    \includegraphics[width = \linewidth]{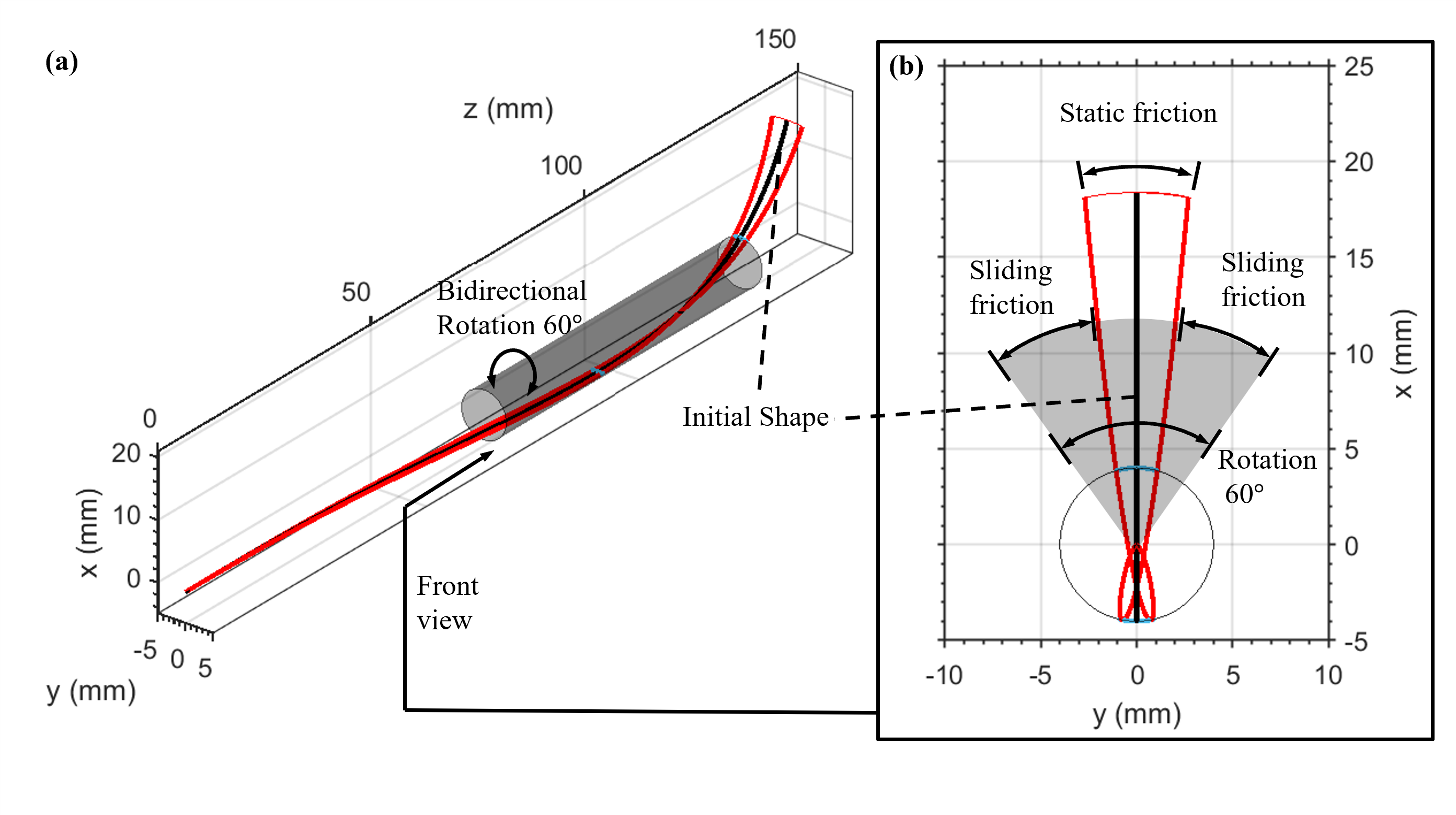}
    \caption{Simulation of a rod in contact with a frictional pipe. (a) The shape trajectory of the rod when rotating the pipe bidirectionally. (b) The front view with the contact phases being annotated.
    }
    \label{fig:simu_case2}
\end{figure}

Contact detection was performed at each time step along the rod centerline, which was uniformly discretized into 300 points. A point was considered to be in active contact when its distance to the outer pipe was smaller than the rod radius (0.66 mm). Neighboring active contact points were treated as redundant (see Figure~\ref{fig:contact_type}), and only the point with the smallest distance to the environment was retained. For each retained contact, the radius offset $\mathbf{r}$ in (\ref{eq:p_T_r}) was determined analytically as the radial vector from the rod centerline point to the point on the rod surface closest to the outer pipe. This contact detection procedure is computationally inexpensive, requiring less than 0.1 ms of CPU time per step on average.
In simulation, the rod was fixed at the base and deflected by the friction as the pipe rotated. As seen in Fig. \ref{fig:simu_case2}(a), the pipe was rotated bidirectionally by 60$\degree$, and the shape of the rod was recorded. The radius of the pipe was 4 mm and a COF of 0.5 was used.

Similar to the rod–plane contact case, the rod switched between the static and sliding friction phases. Initially, the rod was stationary and in contact with the pipe at the top edge and along the body. As the rotation began, the rod was twisted due to static friction and moved along with the pipe at these contact locations. After the friction reached the maximum, the rod entered the sliding friction phase and stopped moving. The $2$ extreme shapes of the rod at the sliding friction phase are shown in the front view in Figure~\ref{fig:simu_case2}(b).

\begin{figure} [h]
    \centering
    \includegraphics[width = 0.9\linewidth]{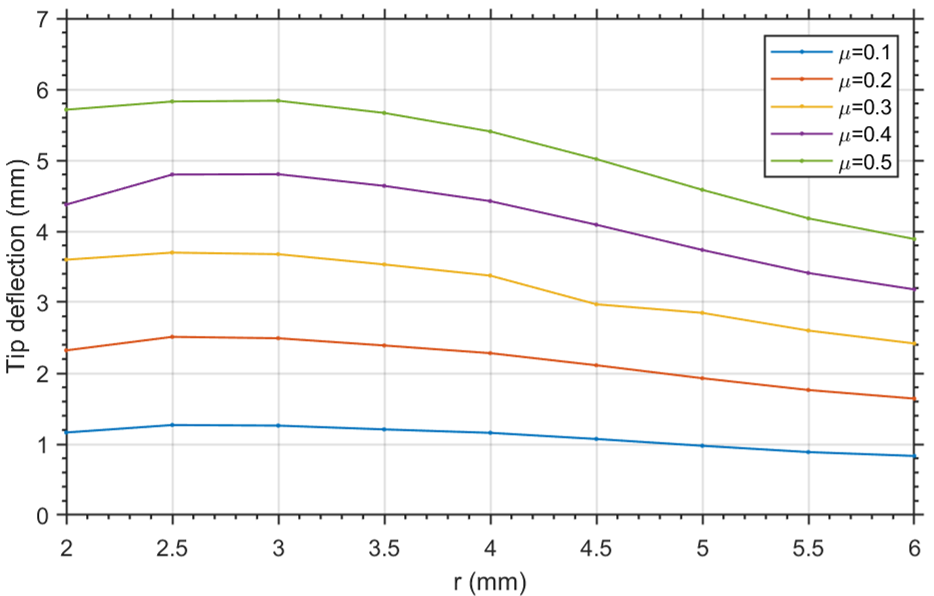}
    \caption{The change of maximum tip deflection with respect to the radius of the pipe and the COF.}
    \label{fig:simu_case2_TABLE}
\end{figure}

The effects of different COFs ($\mu$) and the radius of the pipe ($r$) were studied in Fig. \ref{fig:simu_case2_TABLE}. The maximum tip deflection of the rod was recorded with $\mu \in \{0.1,0.2,\ldots,0.5\}$ and $r \in \{2~\text{mm},2.5~\text{mm},3~\text{mm},\ldots,6~\text{mm}\}$. Since the friction force was proportional to $\mu$, the maximum tip deflection rose linearly as the COF increased. The effect of $r$ was more intricate. As the value of $r$ enlarged, the potential energy of the rod was released and the contact force decreased. On the other hand, the displacement between the two contact points, i.e. the moment arm, also increased, which may lead to a larger total torque about the longitudinal axis and therefore increased maximum tip deflection. 

\subsection{Simulation 3: Rod-pipe-plane Contact} 
\label{sec:simu_case3} 
\begin{figure}[h]
    \centering
    \includegraphics[width = \linewidth]{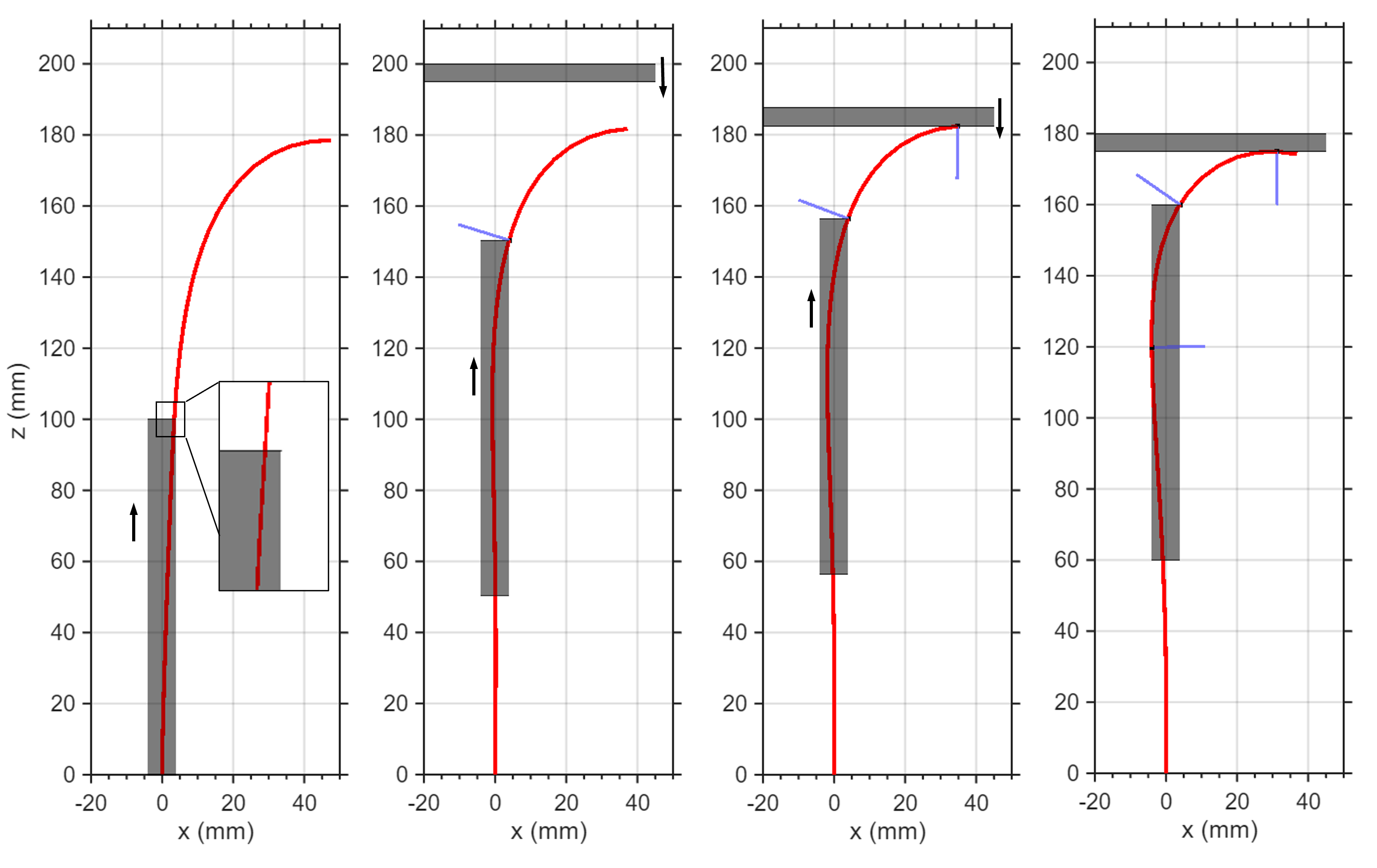}
    \caption{Simulation of a rod simultaneously contacting a plane and a pipe. Sequential frames show the establishment of each contact, with blue arrows indicating the contact force direction.}
    \label{fig:simu_pipeplane}
\end{figure}

\begin{figure}[h]
    \centering
    \includegraphics[width = \linewidth]{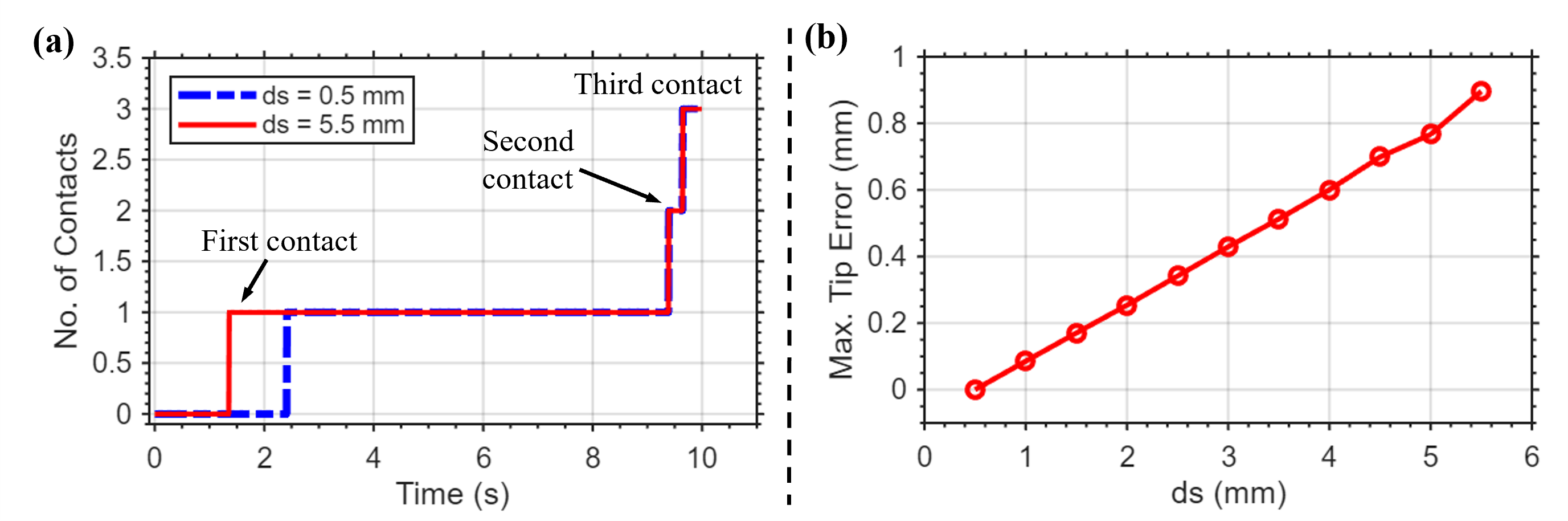}
    \caption{Effect of rod discretization step size on contact behavior and accuracy of the rod-pipe-plane case. (a) Evolution of the number of active contacts over time. (b) Maximum tip error with respect to the reference trajectory.}
    \label{fig:simu_pipeplane_2d}
\end{figure}

\corrlab{R2-2}{
Subsequently, we simulated the rod contacting the plane and the rigid pipe simultaneously. This rod-pipe-plane design is to demonstrate the establishment of corner and surface contacts. The COF of the plane was 0.2, and the parameters of the pipe were kept the same as those in \hyperref[sec:simu_case2]{Simulation 2: Rod-pipe contact}. As shown in Fig. \ref{fig:simu_pipeplane}, the precurved rod is fixed at the base. Starting from a contact-free configuration, the pipe first moved upward to establish the initial corner contact. The plane then moved downward, generating a surface contact and bending the rod downward. Eventually, a third contact was formed between the rod body and the pipe.

We further investigated the sensitivity of the contact detection algorithm to the rod discretization step, defined as $\mathrm{d}s=(s_n-s_1)/(n-1)$. Eleven uniformly spaced values of $\mathrm{d}s$ were tested, ranging from 0.5 to 5.5~mm. The case with $\mathrm{d}s=0.5$~mm was used as the reference trajectory, as it provides the finest discretization. Increasing $\mathrm{d}s$ to 5.5~mm caused the first corner contact to be detected earlier than in the reference case, as shown in Fig.~\ref{fig:simu_pipeplane_2d}(a). In contrast, the second and third surface contacts were detected at approximately the same instants as in the reference trajectory. The earlier detection of the corner contact introduced a maximum tip position error of approximately 1 mm when $\mathrm{d}s=5.5$ mm, and this error increased approximately linearly with $\mathrm{d}s$, as shown in Fig.~\ref{fig:simu_pipeplane_2d}(b). These results illustrate the trade-off associated with increasing the rod discretization step: while a larger $\mathrm{d}s$ improves computational efficiency, it reduces the accuracy of contact detection, particularly for localized corner contacts.
}

\subsection{Simulation 4: CTR-plane Contact }
\label{sec:simu_case4}
\corrlab{R1-4}{
To further demonstrate our method for more complex continuum robots, we simulated a three-tube CTR being actuated to make contact with a frictional plane at the tip. In this case, the ODE of the Cosserat rod formulation in (\ref{eq:ode_cosserat}) has a more complex state vector $\mathbf{y}=[\mathbf{u}_1^T, u_{2,z}, u_{3,z}, \theta_2, \theta_3]^T$ , where $u_{k,z}$ and $\theta_k$ denote the curvature's $z$-component, and the torsion angle of the $k$-th tube, respectively.
}
We refer the readers to our previous work \citet{2024CTR_jia} for the parameters of the CTR and the methods to derive the Jacobian and compliance matrices. The current simulation does not account for internal tube-tube friction in the CTR. 

\begin{figure} 
    \centering
    \includegraphics[width = \linewidth]{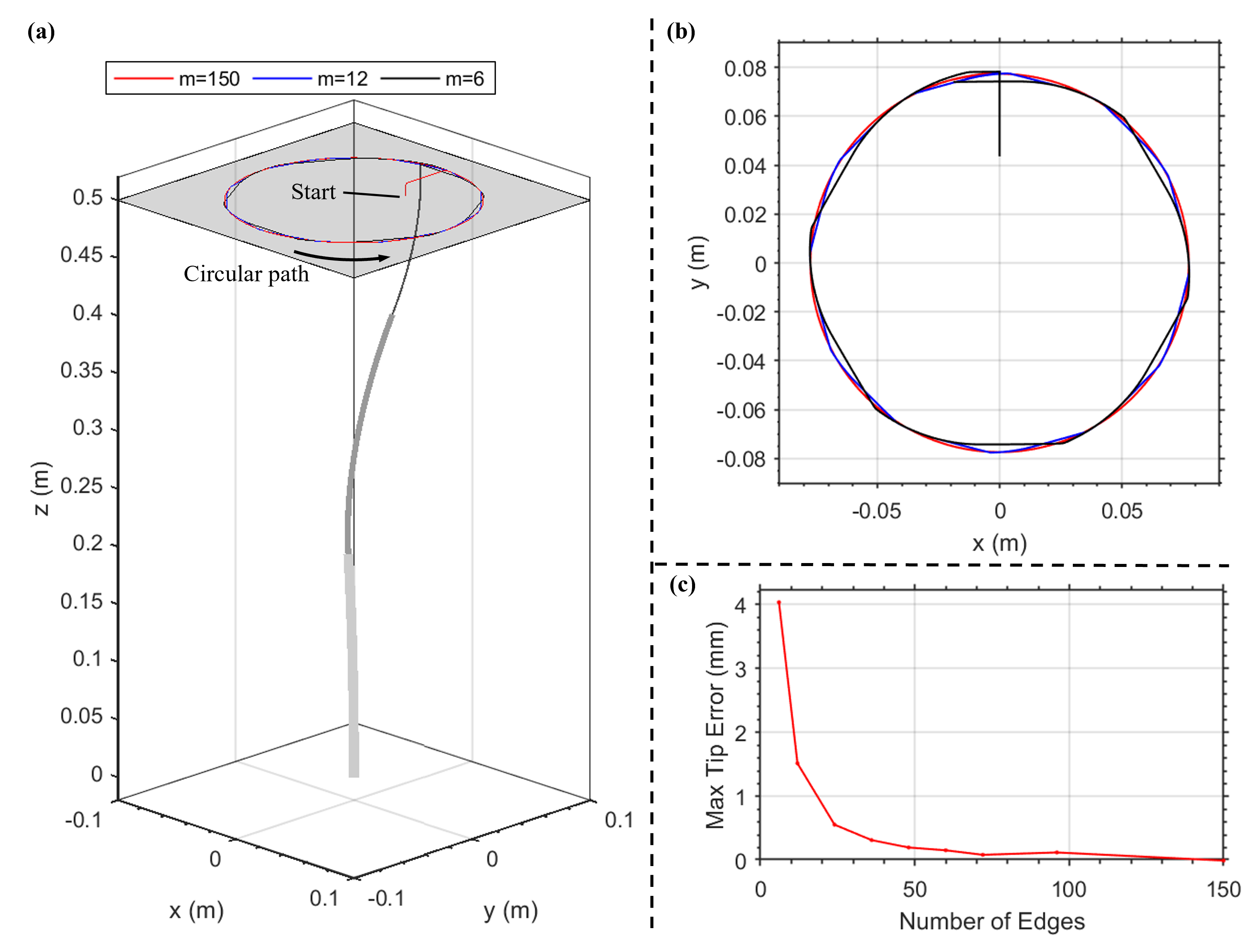}
    \caption{Simulation of a CTR in contact with a frictional plane. The CTR was actuated to contact, slide, and draw a circular path, and the effect of different numbers of edges of the polygonized friction cone ($m$) was studied. (a) The 3D representative plot. (b) The top view of the tip trajectory. (c) The change of maximum tip error w.r.t. $m$.
    }
    \label{fig:simu_ctr}
\end{figure}

As seen in Fig.~\ref{fig:simu_ctr}, the robot first slid on the plane and then drew a circular path. With only the knowledge of the actuation input and the plane's location, the LCP-based model was able to predict contact behaviors and update the robot's shape. We also examined the effect of the friction cone in Figure~\ref{fig:delta_p}(a). The tip trajectories of the CTR were shown in Fig. \ref{fig:simu_ctr}(b) with different numbers of edges of the polygonized friction cone ($m$). We assumed that $m=150$ provided an accurate approximation of the friction cone and used the corresponding tip trajectory as the reference. The COF used in this simulation was 0.3.

Compared with the tip trajectory of $m=150$, using a smaller number of edges ($m=6, 12$) produced a combined trajectory of a circle and polygons. This was attributed to the artifacts of the LCP when the friction cone was linearly approximated. The solution of an LCP maximizes the dissipation of energy, which generates the friction force along the diagonal of the polygonized friction cone rather than the direction of relative sliding ($\mathbf{v}$) \citep{LCP1997_A_P}. When the number of edges ($m$) was large, the error was minimal since the closest diagonal to the direction of $\mathbf{v}$ was selected. As seen in Fig.~\ref{fig:simu_ctr}(c), the maximum tip errors decreased below $0.5$ mm when $m \ge 30$.
We used $m=150$ for better accuracy in the following simulations and experiments.

\subsection{Computational Efficiency}

%

To evaluate the algorithm's computational efficiency, we recorded the average computation time per step for the four simulation scenarios in this section, along with the times required to compute the Jacobian and forward kinematics (FK) and to solve the LCP.
We implemented the algorithm in Matlab on an 8-core 3.00 GHz CPU processor.

As seen in Table \ref{tab:compu_eff}, the total calculation times for all the test scenarios were at most 11.06 ms ($>90$ Hz update rate), which is sufficient for real-time feedback and simulation. The computation of the Jacobian and forward kinematics constituted the primary computational load, whereas solving the LCP was relatively fast, taking less than 1.2 ms per step.

We argue that the overall efficiency of our method can be significantly improved by optimizing the forward kinematics computation. Recent advancements in discretized Cosserat rod models \citep{Nabil_collocation, renda2016discrete} have addressed this by interpolating the curvature function and providing analytical Jacobian and compliance matrices. Furthermore, the calculation speed can be further increased by implementing the algorithm in C++.

\begin{table}[h!]
\renewcommand{\arraystretch}{1.25}
\caption{Computational Efficiency of the LCP-based Model}
\label{tab:compu_eff}
\small\sf\centering
\begin{tabular}{c c c c c}
\toprule
~ & LCP Size & Total time & Jacob. & LCP  \\
~ & - & [ms] & [ms] & [ms] \\
\midrule
Rod-plane      &   152 & 6.96  &  3.89 &  0.25 \\
Rod-pipe       &   304 & 7.78  &  4.64 &  0.76 \\
Rod-pipe-plane &   456 & 8.70  &  4.99 &  1.14  \\
CTR-plane      &   152 & 11.06 &  8.88 &  0.31  \\
\bottomrule
\end{tabular}
\vspace{- 2mm}
\end{table}

\section{Experimental Results} \label{sec:experiments}

To validate the proposed friction model, we conducted physical experiments on two setups: 1) A single tube interacting with complex environments. Three sets of experiments were conducted for this setup. The first experiment considered a single surface contact between the tube and a rigid plane. On top of that, another circular ring was added to the environment in the second experiment to validate the model in the presence of both surface and corner contact. The last experiment tested the model with the tube inside a rigid straight pipe, which resembles a two-tube CTR with a large clearance. In these three scenarios, we moved the environment with the tube fixed at the base to manifest the effect of friction. 2) A tendon-driven continuum robot (TDCR) interacting with a rigid wall. This setup validated the accuracy of the model in an actuated robot.
The animations of these experimental results were shown in \hyperref[appendix]{Extension 1}.


The accuracy of the proposed model was evaluated from four performance indices that compared the measured trajectory of the tube with the simulation:

\noindent 1) The maximum error of the tip position:
\begin{equation}
    \epsilon_{\text{tip}} = \max_{j=1,\ldots,N}  \lVert \mathbf{p}_{\text{tip},j}^{\text{model}} - \mathbf{p}_{\text{tip},j}^{\text{exp}} \rVert.
\end{equation}
2) The mean error of the body shape along the trajectory:
\begin{equation}
    \epsilon_{\text{body}} = \frac{1}{N} \sum_{j=1}^N \left ( \frac{1}{n} \sum_{i=1}^n \lVert \mathbf{p}^{\text{model}}_j(s_i) - \mathbf{p}_j^{\text{exp}} (s_i)\rVert \right).
\end{equation}
3) The maximum error of the body shape: 
\begin{equation}
    \epsilon_{\text{max}} = \max_{i,j}  \lVert \mathbf{p}_j^{\text{model}}(s_i) - \mathbf{p}_j^{\text{exp}}(s_i) \rVert.
\end{equation}
4) The error of the maximum tip deflection: 
\begin{equation}
    \epsilon_{\text{d}} =  | d_{\text{tip}}^{\text{model}} - d_{\text{tip}}^{\text{exp}} |,
\end{equation}
where $d_{\text{tip}}^{(\cdot)}$ denotes the maximum tip deflection:
\begin{equation}
    d_{\text{tip}}^{(\cdot)} = \max_{j_1,j_2 \in \{1,\ldots,N\}} \lVert \mathbf{p}_{\text{tip},j_1}^{(\cdot)} - \mathbf{p}_{\text{tip},j_2}^{(\cdot)} \rVert.
\end{equation}
We remark that the superscripts `model' and `exp' are used to represent the model predictions and experimental measurements, respectively. The numbers of discretized arc-length points and time steps are denoted as $n$ and $N$, respectively, with $i$ and $j$ being the corresponding arc-length and time indices. We further note that these metrics do not explicitly quantify torsional deformation, since no significant pure torsional deformation was observed in the experimental setup.


\subsection{Tube-plane Experiment} \label{sec:tube_plane}
\begin{figure} 
    \centering
    \includegraphics[width = \linewidth]{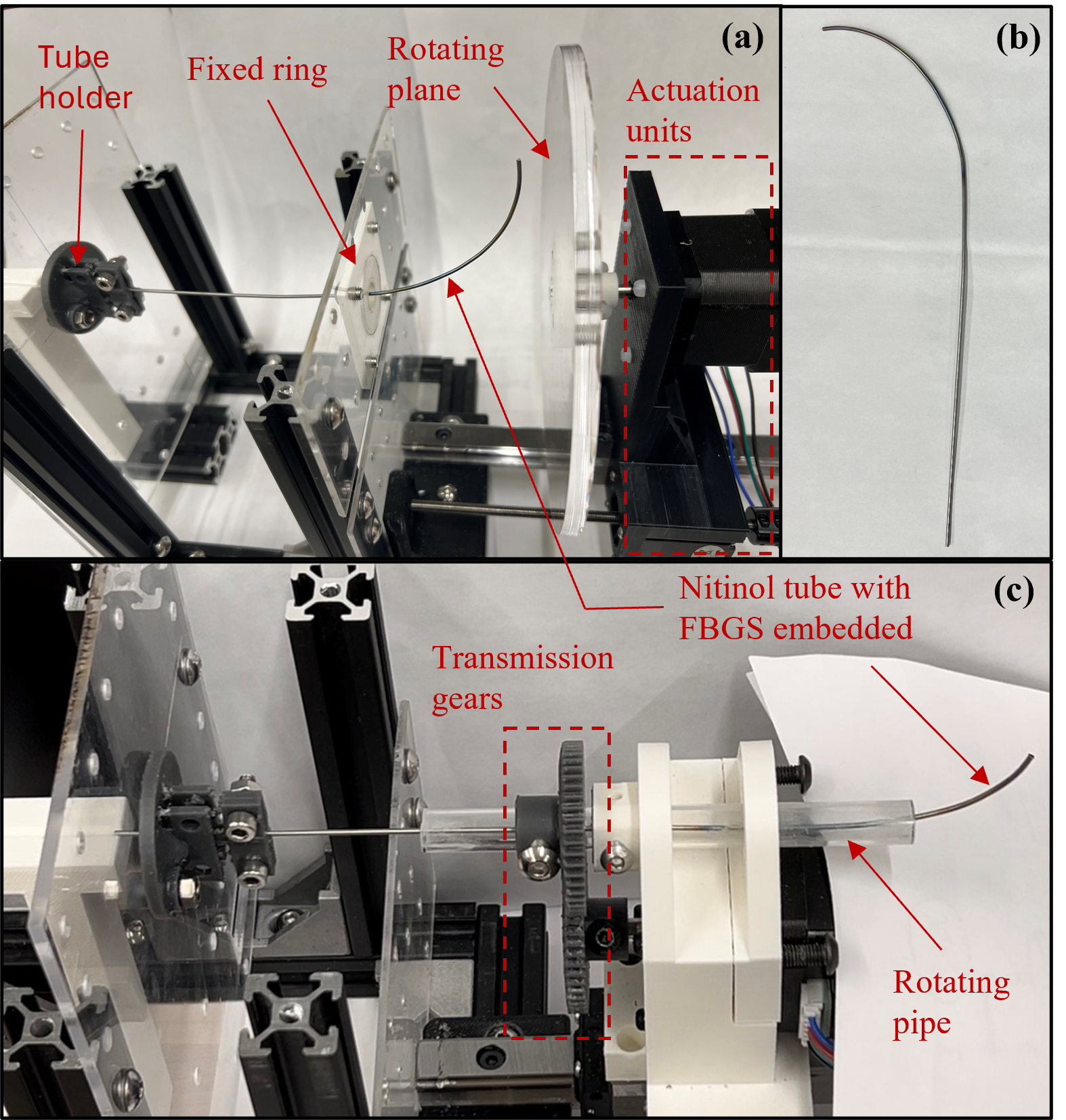}
    \caption{The setup of physical experiments where the tube is fixed at the base and its shape is changed by friction due to the rotation of the environment. (a) Tube-ring-plane scenario: the tube is compressed by a plane while being constrained inside a ring. (b) The precurved nitinol tube used in the experiments. (c) Tube-pipe scenario: the tube is constrained inside a rotating straight pipe.}
    \label{fig:exp_setup}
\end{figure}

Firstly, we validated the model for a single-contact scenario where the tube is compressed by a laser-cut acrylic disk, as seen in Fig. \ref{fig:exp_setup}(a). The acrylic disk was actuated by stepper motors to enable the insertion and rotation. The position and orientation of the tube base were fixed by the holder, and its shape was changed by moving the acrylic disk. 
The straight part of the tube was designed to be collinear with the center of the disk. 
Note that the circular ring was not yet added in the first experiment. 
The visibility of the whole shape was limited because of the contact environments and transmission mechanisms, which precluded the use of vision-based sensing methods. 
Therefore, we embedded the Fiber Bragg Grating Sensor (FBGS) inside the lumen of the tube (Fig. \ref{fig:exp_setup}(b)) and fixed it at both the base and the tip using tape to restrict relative torsion. The FBG sensor provided sensing points spaced 20 mm apart along the arc length, which measured the local curvature and then reconstructed an interpolated 3D centerline of the tube. The calibration and shape reconstruction were implemented in \textit{ILLumiSense} and \textit{Shape Sensing} Software. The accuracy of the FBG sensor has been verified in our previous work \citep{Previous_work} with the reported mean body error of $1.11$ mm. 

We assumed that the frictional effect from the translation between the tube and the environment was negligible due to the high rigidity along the longitudinal axis. This assumption has been reported in \citet{frictionCTR2011} for the CTR, and we will further validate it for a large clearance tube-pipe pair scenario in the \hyperref[sec:tube_pipe]{Tube-pipe experiment}. Then the contact model was validated by itself without friction by inserting the disk toward the tube. 
Consistent with the assumption of no initial contact, the plane was initially positioned away from the tube such that no contact existed. It was then moved toward the tube until sufficient contact force was established for the friction analysis.
The tube shape was subsequently recorded at 5~mm insertion intervals and compared with the simulation results.
For each insertion length, a 90-degree bidirectional rotation of the plane was conducted and the rotations were performed three times continuously to ensure the repeatability.
In the simulations, we used a step size of 1 mm for insertion and \(1^\circ\) for rotation, which provided sufficient accuracy for the present setup.
Since friction is related to the sliding velocity between the contact objects, two different speeds were used for each rotation, namely 5$\degree$/s and 10$\degree$/s, resulting in 10 groups of experiments in total. 
The performance of our model was compared with two other models: 1) the sliding friction
model, which assumes sliding friction with a direction opposite to the direction of actuation. This is a representative method to approximate the friction without considering the stick motion \citep{Nabil_friction2024, ctr_friction_ha2019, TDCR_stability2023}; 2) frictionless model (FLM), which finds the external contact by minimizing the potential energy with geometric constraint \citep{2024CTR_jia}.

\begin{figure*}
    \centering
    \includegraphics[width = \linewidth]{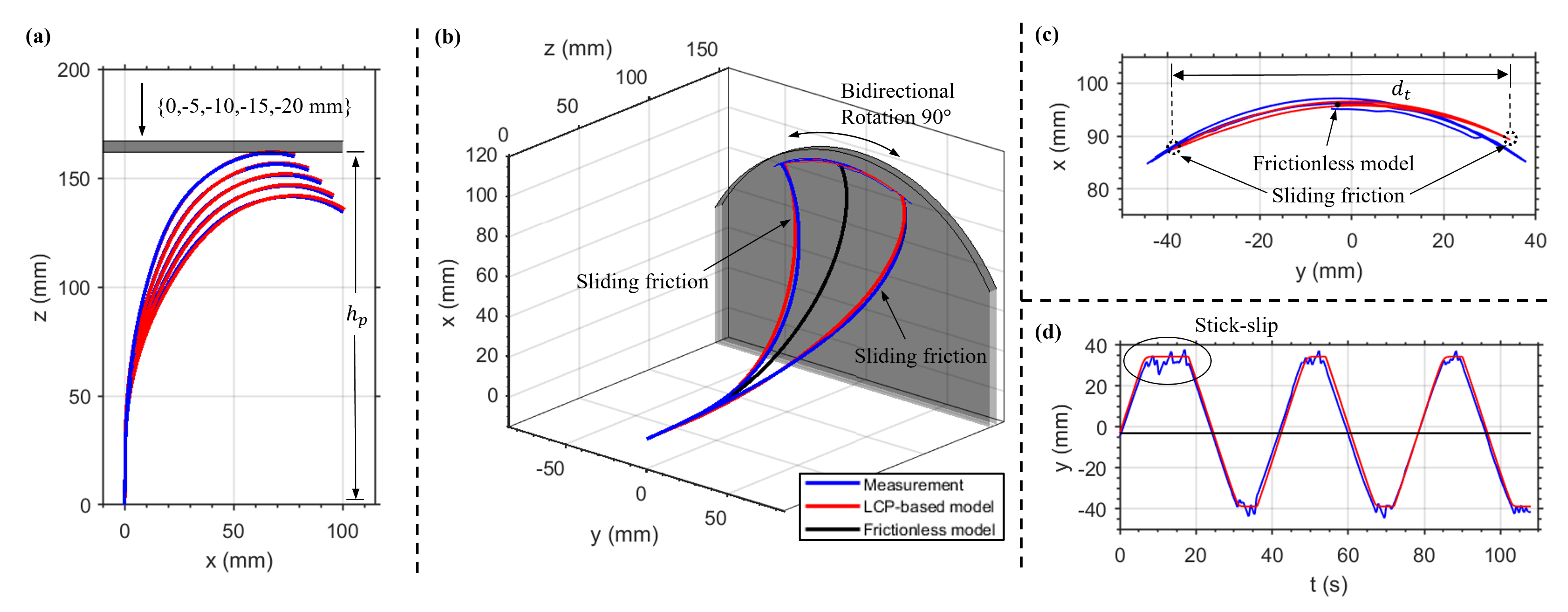}
    \caption{Representative experimental results compared with friction and frictionless model for the tube-plane scenario: (a) Different bending shapes of the tube when compressing the plane in a 5 mm sequence. (b) The 3D trajectory of the tube under 90-degree bidirectional rotation of the plane. The extreme tube shapes are shown at the instants when the rotation direction is reversed. For clarity, the prediction of the sliding friction model is not plotted separately, since it overlaps with the LCP-based model at these boundary configurations. (c) The trajectory of the tip position under the bidirectional rotation. (d) The change of the $y$-axis component of the tip position under the bidirectional rotation. The vibration of the trajectory due to the stick-slip effect was annotated.}
    \label{fig:exp_1tube}
\end{figure*}

The collected dataset was divided into two subsets. The first subset consisting of two insertion lengths \{0 mm, 20 mm\} was used for calibration ($\times$2 different rotation speeds), and the remaining 6 groups were the test dataset. In this experiment, we minimize the sum of tip position, body shape, and maximum deflection error over the COF of the plane $\mu_p$ and the Poisson's ratio $\nu$ of the tube.
\begin{equation} \label{eq:calib}
        \min_{\mu_p, \nu} \sum_{x=1}^{n_{\text{calib}}} \left(\epsilon_{\text{tip},x} + \epsilon_{\text{body},x} + \epsilon_{\text{d},x}\right)
\end{equation}
where $n_{\text{calib}}$ denotes the number of data in the calibration dataset. The Nelder–Meade simplex algorithm was used for the optimization, as implemented by MATLAB’s \textit{fminsearch} function. The calibrated parameters are shown in Table \ref{calib_param} alongside the tube parameters, where $K_b$ is the bending stiffness, and $\mu_r$ and $\mu_c$ are to be introduced in the following experiments. We remark that the tube's bending stiffness is excluded from calibration because its absolute scale does not affect the predicted shape in these rigid-contact experiments. Its assigned value is therefore a scaling parameter, not a measured physical stiffness. 
The calibrated Poisson’s ratio is 0.28, which is close to the nominal value of approximately 0.30 reported for Nitinol.
The distance from the tube base to the plane, denoted as $h_p$ in Fig. \ref{fig:exp_1tube}(a), was manually measured as 160.5 mm for 0 mm insertion. 

\begin{table}[h!]
\renewcommand{\arraystretch}{1.25}
\caption{Parameters Used in the Friction Model }
\label{calib_param}
\small\sf\centering
\begin{tabularx}{0.49\textwidth}{c p{0.151\linewidth}  p{0.085\linewidth} c c c c}
\toprule
ID \textbackslash \, OD & Total Len. & ~$K_b$ & $\nu$ & $\mu_p$ & $\mu_r$ & $\mu_c$ \\
$[\text{mm}]$ & $[\text{mm}]$ & $[\text{N mm}^2]$  & -- & -- & -- & -- \\
\midrule
0.91 \textbackslash \, 1.32 & 202.88 & 20.07 & 0.28 & 0.14 & 0.05 & 0.37\\
\bottomrule
\end{tabularx}
\vspace{- 2mm}
\end{table}

As seen in Fig.~\ref{fig:exp_1tube}(a), the bending shapes of the tube at different insertion lengths agreed with the experiment, which validated the predictions of contact location and forces in the proposed model. The 3D representation of the bidirectional rotation experiment at 10$\degree$/s and $h_p=155.5$ mm was depicted in Fig. \ref{fig:exp_1tube}(b) with the corresponding change of the radial ($y$-axis) component of the tip position in Fig.~\ref{fig:exp_1tube}(d). 

Initially, the tube made contact with the plane at the body and the contact was in the static friction phase. As the rotation began, the tube was twisted due to friction and moved along with the plane without sliding. In this stage, the $y$-axis component of the tip position changed linearly (Fig. \ref{fig:exp_1tube}(d)) since the constant rotation speed was used in the experiment. After reaching the boundary of the friction cone, the friction force and the corresponding tube twist were saturated, and the tube was left behind the rotating plane. Therefore, the tube-plane contact ran into the sliding friction phase where the relative displacement $\mathbf{v}$ was non-zero. The stick-slip phenomenon was observed in the sliding friction phase, as the motion of the objects was typically irregular, characterized by brief accelerations interspersed with periods of rest (Fig. \ref{fig:exp_1tube}(d)). 
We note that additional effects such as the Stribeck effect, where friction decreases with increasing sliding velocity at low speeds, are not explicitly modeled and may contribute to the observed stick–slip behavior.
When the rotation direction switched, the relative motion tendency changed and the tube-plane contact returned to the static friction phase. The motion of the tube switched between the static and sliding friction phases until the actuation ended. 
Our model accurately captured the transition from static to sliding friction through the complementarity constraints. In contrast, the sliding-friction model only predicted the two extreme configurations corresponding to the maximum frictional effect, as shown in Fig.~\ref{fig:exp_1tube}(b-c). This suggests that the sliding friction model adopted in many existing approaches \cite{Nabil_friction2024, TDCR_stability2023, ctr_friction_ha2019} captures only the boundary behavior of friction and cannot reproduce the intermediate phase transition when the actuation direction is reversed. For clarity, the prediction of the sliding-only model is not shown separately in Fig.~\ref{fig:exp_1tube}, since it coincides with our model at the corresponding boundary configurations.

As seen in Fig. \ref{fig:exp_1tube}(b), the predicted shape of the tube at the sliding friction phase and trajectory of the tip position aligned with the experiment. Consequently, it indicated that the proposed model was capable of capturing the switching between the two contact phases. The main error came from the unmodeled stick-slip phenomenon as the maximum static friction always exceeds the sliding friction force. However, the proposed model used a constant COF $\mu$ in (\ref{eq:CCs_friction2}) based on Coulomb’s friction law, and only predicted a single shape of the tube corresponding to $\mu$. The maximum reported error of the tip position was 3.07 mm for the proposed model, which was not significant compared with the larger than 80 mm deflection of the tip position along the trajectory in Fig. \ref{fig:exp_1tube}(c). On the other hand, the frictionless model failed to account for any of these effects and resulted in large modeling errors.

The calculated error metrics for the remaining data were reported in Table \ref{table_exp1}. Over all tube-plane experiments, the proposed friction model predicted the trajectory of the tube with a mean body error of 2.69 mm and a maximum body error of 6.64 mm. The mean and maximum body errors are approximately 1.3\% and 3.3\% of the total tube length, respectively. For the tube-plane setup, the maximum reported error was reasonable since the tube could slip easily on the surface. In applications, the tube is often placed inside other environments, for instance, the outer tube of CTR or the trocar, which enforces a convex geometry constraint to the tube and the stick-slip motion can be mitigated. In the following sections, we conducted experiments with the tube placed inside a fixed ring or a rotating pipe to validate this assumption.


\begin{table}[h!]
\renewcommand{\arraystretch}{1.25}
\caption{Error Analysis for the Tube-plane Experiment}
\label{table_exp1}
\small\sf\centering
\begin{tabularx}{0.37\textwidth}{c c p{0.07\linewidth} p{0.07\linewidth} p{0.07\linewidth} p{0.07\linewidth}}
\toprule
$h_p$ & Speed  & $\epsilon_{\text{tip}}$  & $\epsilon_{\text{body}}$ & $\epsilon_{\text{max}}$ & $\epsilon_{\text{d}}$ \\
$[\text{mm}]$ & $[\degree/s]$ & $[\text{mm}]$  & $[\text{mm}]$ & $[\text{mm}]$ & $[\text{mm}]$ \\
\midrule 
155.5  &  5 & 3.78 & 2.52 & 5.79 & 3.93 \\
155.5  &  10 & 3.07 & 2.53 & 5.61 & 0.15 \\
150.5 & 5 & 2.41 & 2.74 & 	5.66 & 	2.80 \\
150.5 & 10 & 5.47 & 2.69 &	6.52 & 	4.99 \\
145.5  &  5 & 3.08 & 2.67 & 5.44 & 0.38 \\
145.5  &  10 & 3.92 & 3.03 & 6.64 & 5.36 \\
\bottomrule
\end{tabularx}
\end{table}

\subsection{Tube-ring-plane Experiment} \label{sec:tube_ring_plane}

Subsequently, we conducted the second experiment with the ring added to constrain the tube as seen in Fig. \ref{fig:exp_setup}(a). The ring was 3D printed and fixed to the environment. The concentricity among the tube, ring, and disk was ensured during assembly. In this setup, the effect of frictional contact was significant since as the disk pressed the tube down, the ring pushed against the tube and generated a large contact force. 

Analogously, we inserted the plane with a sequence of \{0, 5, 10, 15 mm\} and rotated the plane 90-degree bidirectionally with the two speeds, which resulted in 8 groups of data. Then the insertion groups \{5, 15 mm\} ($\times$2 rotation speeds) were used to calibrate the COF of the ring $\mu_r$ by minimizing (\ref{eq:calib}), and the remaining 4 groups of data were for error analysis. The calibrated $\mu_r$ was 0.05, as reported in Table \ref{calib_param}. The small value of $\mu_r$ matched our physical intuition since the ring was 3D-printed with smooth edges. In this scenario, the measured distance $h_p$ for 0 mm insertion was also 160.5 mm.

\begin{figure*}
    \centering
    \includegraphics[width = \linewidth]{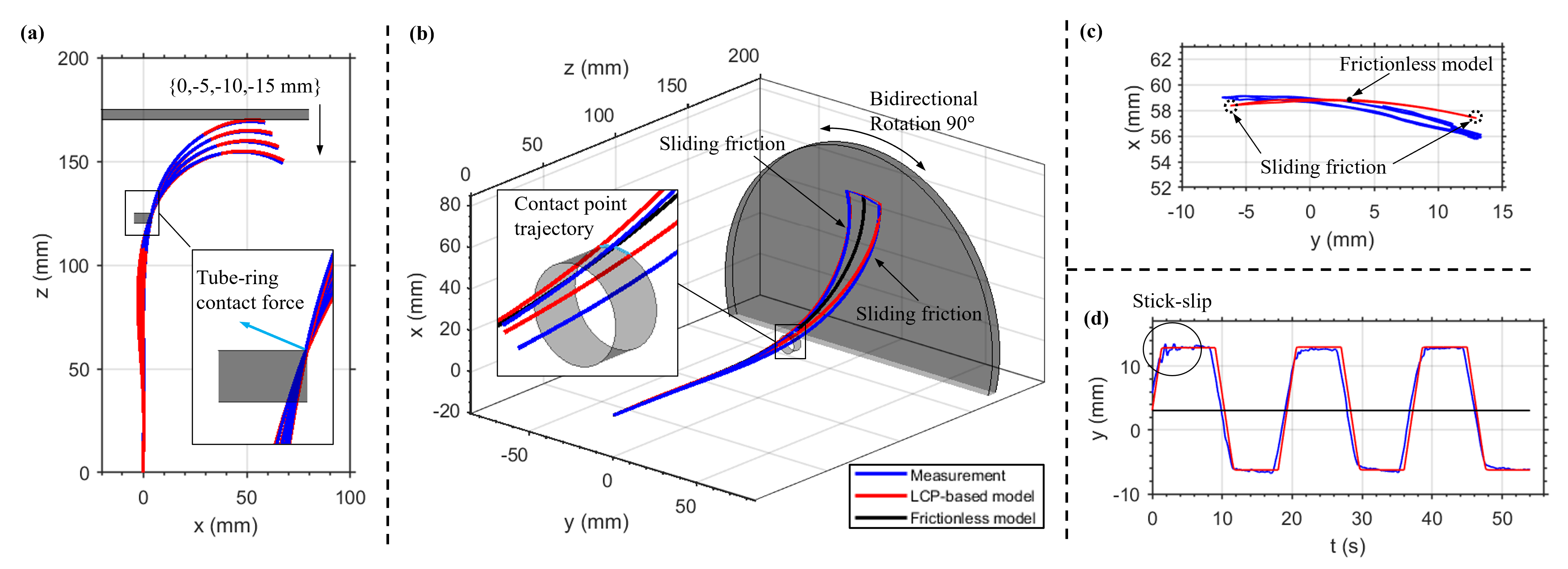}
    \caption{Representative experimental results compared with friction and frictionless model for the tube-ring-plane scenario: (a) Different bending shapes of the tube when compressing the plane in a 5 mm sequence. The corner contact force at 15 mm insertion is shown. (b) The 3D trajectory of the tube under 90-degree bidirectional rotation of the plane. The extreme shapes of the tube were shown when the direction of rotation shifted. A zoom-in plot was used to demonstrate the simulated trajectory of the contact point at the ring. (c) The trajectory of the tip position under the bidirectional rotation. (d) The change of the $y$-axis component of the tip position under the bidirectional rotation. The vibration of the trajectory due to the stick-slip effect was annotated.}
    \label{fig:exp_2contact}
\end{figure*}

The bending shapes of the tube were depicted in Fig.\ref{fig:exp_2contact}(a), and the tube-ring contact force at 15 mm insertion was shown in the zoom-in plot. The proposed friction model accurately estimated the shape of the tube under the 2 contacts, with the direction of the tube-ring contact force nearly normal to the tube surface. 

A representative experiment with $h_p = 160.5$ mm and 10$\degree$/s rotation speed was demonstrated in Fig. \ref{fig:exp_2contact}(b-d). Likewise, the trajectory of the tube experienced the static friction phase and the sliding friction phase as described in the \hyperref[sec:tube_plane]{Tube-plane experiment} section. The proposed model was able to capture both contact phases and their transition. The simulated trajectory of the tube-ring contact point was depicted in the zoom-in plot, which matched physical insights that the tube always made contact with the top edge of the ring. Note that an offset of 1.5 mm was found between the measured tube shape and the edge of the ring. This might be caused by the drift of FBGS or the slight misalignment of the ring during assembly. Despite this shape error at the ring, the shape of the tube was accurately predicted with a mean body error of 0.93 mm along the trajectory. The stick-slip phenomenon was still observed, as seen in Fig.\ref{fig:exp_2contact}(d), but less significant compared with the tube-plane scenario in Fig.\ref{fig:exp_1tube}(d).

 The error metrics of this experiment were analyzed in Table \ref{table_exp2}. The maximum tip error, mean body error, and maximum body error were 2.05 mm, 1.06 mm, and 2.51 mm, respectively, across all groups of test data. Additionally, we compared the error at $h_p=150.5$ mm with that reported in the \hyperref[sec:tube_plane]{Tube-plane experiment} section at the same height. With the constraint of the ring, the maximum body error induced by the stick-slip motion was reduced from 6.52 mm (Table \ref{table_exp1}) to 2.51 mm by 61.5\%. 
The ring acts as an additional geometric constraint that modifies the normal loading and contact distribution along the tube, while also limiting the deformation range. Although these effects may mitigate stick–slip motion, the present comparison does not isolate that mechanism from the reduced deformation severity. Therefore, the observed improvement is hypothesized to result from the combined effects of modified normal loading, redistributed contact, and constrained deformation.

\begin{table}[h!]
\renewcommand{\arraystretch}{1.25}
\caption{Error Analysis for the Tube-ring-plane Experiment}
\label{table_exp2}
\small\sf\centering
\begin{tabularx}{0.37\textwidth}{c c p{0.07\linewidth} p{0.07\linewidth} p{0.07\linewidth} p{0.07\linewidth}}
\toprule
$h_p$ & Speed  & $\epsilon_{\text{tip}}$  & $\epsilon_{\text{body}}$ & $\epsilon_{\text{max}}$ & $\epsilon_{\text{d}}$ \\
$[\text{mm}]$ & $[\degree/s]$ & $[\text{mm}]$  & $[\text{mm}]$ & $[\text{mm}]$ & $[\text{mm}]$ \\
\midrule 
160.5  &  5 & 1.82	 & 1.05	& 2.30  & 3.08\\
160.5  &  10 & 1.19 & 0.93	& 1.92	& 1.87 \\
150.5  &  5  & 2.05 & 1.13	& 2.19	& 0.08\\
150.5  &  10 & 1.94 & 1.13	& 2.51	& 3.14\\
\bottomrule
\end{tabularx}
\end{table}

\subsection{Tube-pipe Experiment} \label{sec:tube_pipe}
\begin{figure}
    \centering
    \includegraphics[width = \linewidth]{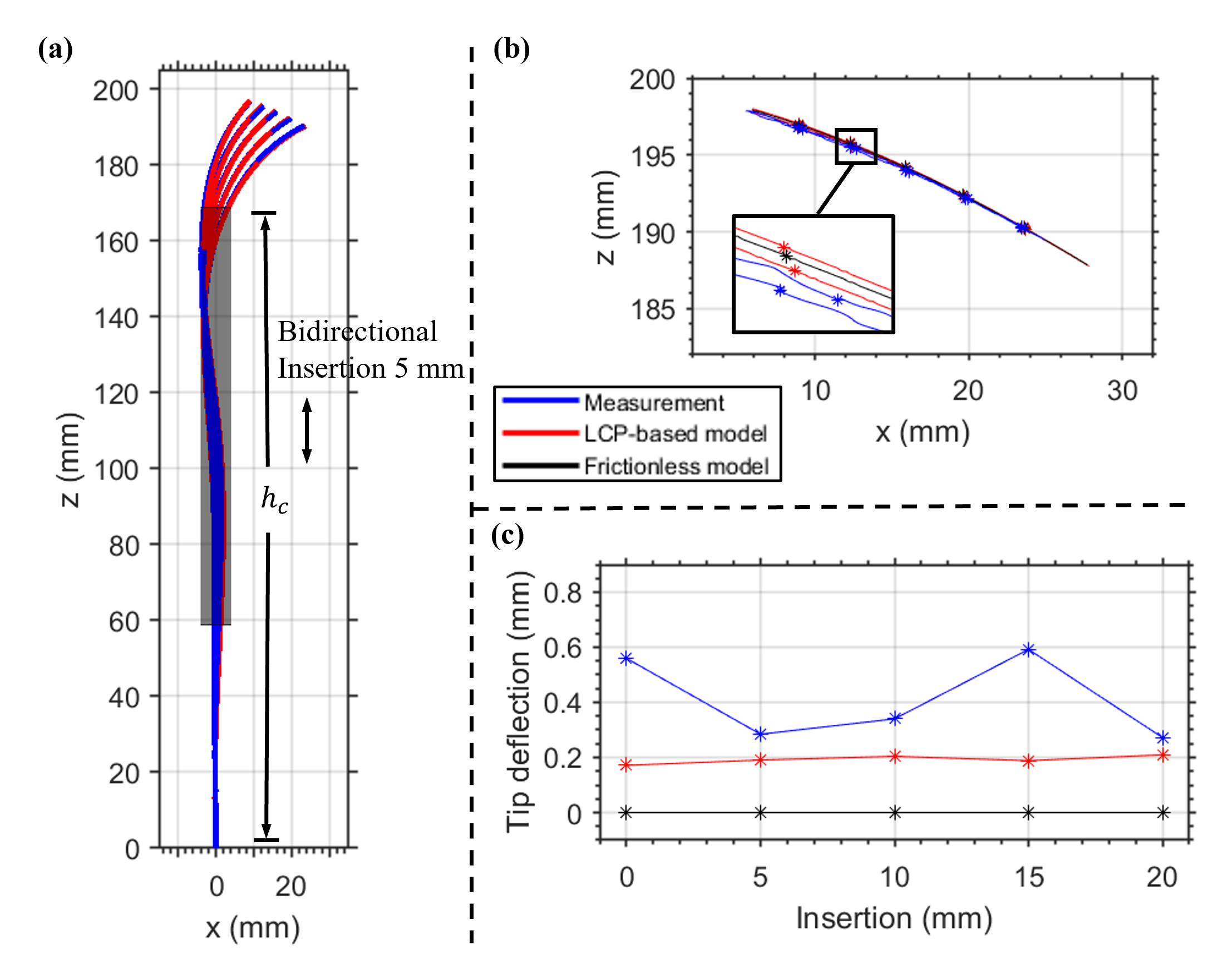}
    \caption{Experimental results compared with the friction model for the tube-pipe scenario when the pipe was translated bidirectionally in a 5 mm sequence: (a) Different bending shapes of the tube. (b) The trajectory of the tip position under the bidirectional translation. (c) Deflection of the tip position for the same insertion length under different directions.
    }
    \label{fig:exp_cylinder_insert}
\end{figure}

In the third experiment, the tube was placed inside a straight acrylic pipe as shown in Fig. \ref{fig:exp_setup}(c). A pair of transmission gears was used to enable the rotation of the pipe. This setup resembled the 2-tube CTR with a straight outer tube and large clearance. The distance from the tube base to the top of the pipe was defined as $h_c$ in Fig. \ref{fig:exp_cylinder_insert}(a). The tube would make contact with the pipe at the top edge and along the inner surface, and we studied the shape change of the tube under the interaction between these contacts.

We first investigated the effect of friction under bidirectional translation of the pipe. 
The shapes of the tube were recorded at five different values of $h_c$ at every 5 mm between  163.8 mm and 183.8 mm.
At each distance, shapes obtained from both forward and backward insertion were collected and compared with the simulation. The COF used in the simulation was 0.37 at the translation stage. Fig. \ref{fig:exp_cylinder_insert}(a-b) depicted the tube shape at different $h_c$ and the corresponding tip positions. No evident shape discrepancies were found between forward and backward insertion and the shapes superposed with each other. The tube shapes from the frictionless model were not shown in Fig. \ref{fig:exp_cylinder_insert}(a) for better visualization. As shown in Fig. \ref{fig:exp_cylinder_insert}(c), the maximum deflection of tip position for equal insertion lengths with varying insertion directions was 0.6 mm in the experiment, and 0.2 mm in the friction model. We remark that a similar observation has been reported in \citet{frictionCTR2011} that the effect of friction was minimal when the tubes of a CTR were translated.
This was caused by the longitudinal rigidity of the tube compared to the torsional compliance. The friction induced by the translation lies in the tube's longitudinal direction and leads to smaller shape deflection compared with that in the torsional direction.

The effect of the friction was also studied by bidirectionally rotating the pipe by $45\degree$.
The pipe was inserted into the heights at every 5 mm between 163.8 mm and 188.8 mm and two different rotating speeds were used for each experiment. The data at the heights of \{$163.8$ mm, $173.8$ mm, $183.8$ mm\} were used for calibrating the COF of the pipe, $\mu_c$. By optimizing (\ref{eq:calib}), we obtained $\mu_c=0.37$.

\begin{figure*}
    \centering
    \includegraphics[width = \linewidth]{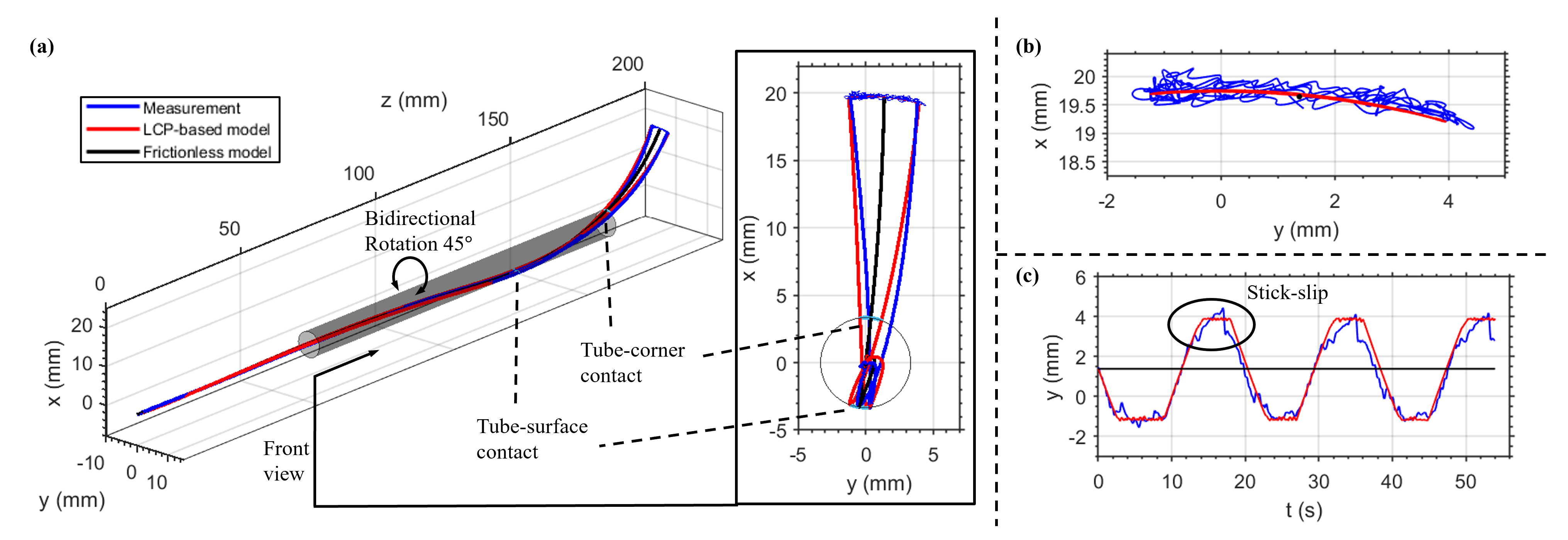}
    \caption{Representative experimental results compared with friction and frictionless model for the tube-pipe scenario when the pipe was rotated bidirectionally with 45 degrees: (a) The 3D trajectory of the tube. The extreme shapes of the tube were shown when the direction of rotation shifted. The front view was presented with the simulated trajectories of the contact points along the pipe. (b) The trajectory of the tip position under the bidirectional rotation. (c) The change of the $y$-axis component of the tip position under the bidirectional rotation. The vibration of the trajectory due to the stick-slip effect was annotated.}
    \label{fig:exp_cylinder_rot}
\end{figure*}

We selected a representative experiment with $h_c = 178.8 $ mm and 5$\degree$/s rotation speed and the 3D shape trajectory was shown in Fig. \ref{fig:exp_cylinder_rot}(a). The tube-pipe contacts switched between the static and sliding friction phases in a similar manner described in the \hyperref[sec:tube_plane]{Tube-plane experiment} section, which was consistent with the simulation. The front view was depicted with the trajectories of the contact points annotated. The tube made two contacts with the pipe, one at the top edge (corner contact) and the other along the inner surface of the pipe (surface contact). Despite a 1 mm offset between the contact trajectories, both contacts were correctly modeled with negligible penetrations.

As seen in Fig. \ref{fig:exp_cylinder_rot}(b-c), the stick-slip phenomenon was more pronounced in this experiment compared with tube-plane or tube-ring-plane scenarios, and the trajectory of the tip position contained more vibrations. Two main reasons for the increased stick-slip phenomenon were speculated. Firstly, the COF of the pipe $\mu_c$ was much larger than $\mu_p$ (Table \ref{calib_param}) so the effect of friction was more significant. At the same time, the corner contact and the surface contact interfered with each other and exacerbated the stick-slip motion. For example, when the surface contact reached the sliding friction phase, the vibration near the contact region spread to the whole shape, destabilizing the corner contact. 
When both contacts reached the sliding friction phase, their vibration accumulated, resulting in an even larger stick-slip motion.

The error metrics for the tube-pipe experiment under bidirectional rotation were presented in Table \ref{table_exp3_rot}. In all experiments, the mean body error and maximum body error were 0.55 mm and 2.83 mm, respectively. The maximum error of the tip position due to the stick-slip motion was 1.26 mm, equivalent to 16.1\% of the maximum tip deflection (7.79 mm). Since the frictionless model did not account for any tip deflection, we argue that the proposed friction method significantly reduced the model inaccuracy by 83.8\%. 
\corrlab{R1-3}{
We remark that the FBG reconstruction reported a mean body error of 1.11 mm. Therefore, the sub-millimeter errors in Table \ref{table_exp3_rot} indicate that the model prediction was within the uncertainty of the FBG-reconstructed reference, rather than demonstrating independently verified accuracy. 
}

\begin{table}[h!]
\renewcommand{\arraystretch}{1.25}
\caption{Error Analysis for the Tube-pipe Experiment (Rotation)}
\label{table_exp3_rot}
\small\sf\centering
\begin{tabularx}{0.37\textwidth}{c c p{0.07\linewidth} p{0.07\linewidth} p{0.07\linewidth} p{0.07\linewidth}}
\toprule
$h_c$ & Speed  & $\epsilon_{\text{tip}}$  & $\epsilon_{\text{body}}$ & $\epsilon_{\text{max}}$ & $\epsilon_{\text{d}}$ \\
$[\text{mm}]$ & $[\degree/s]$ & $[\text{mm}]$  & $[\text{mm}]$ & $[\text{mm}]$ & $[\text{mm}]$ \\
\midrule 
168.8  &  5 & 0.82 & 0.41 & 0.86 & 1.06 \\
168.8  &  10 & 0.56 & 0.48 & 2.83 & 1.45 \\
178.8  &  5 & 0.98 & 0.55 & 1.47 & 0.03 \\
178.8  &  10 & 0.76 & 0.53 & 1.21 & 1.10 \\
188.8  &  5 & 1.01 & 0.72 & 1.33 & 0.02 \\
188.8  &  10 & 1.26 & 0.75 & 1.65 & 0.77 \\
\bottomrule
\end{tabularx}
\end{table}


\begin{figure*}
    \centering
    \includegraphics[width = \linewidth]{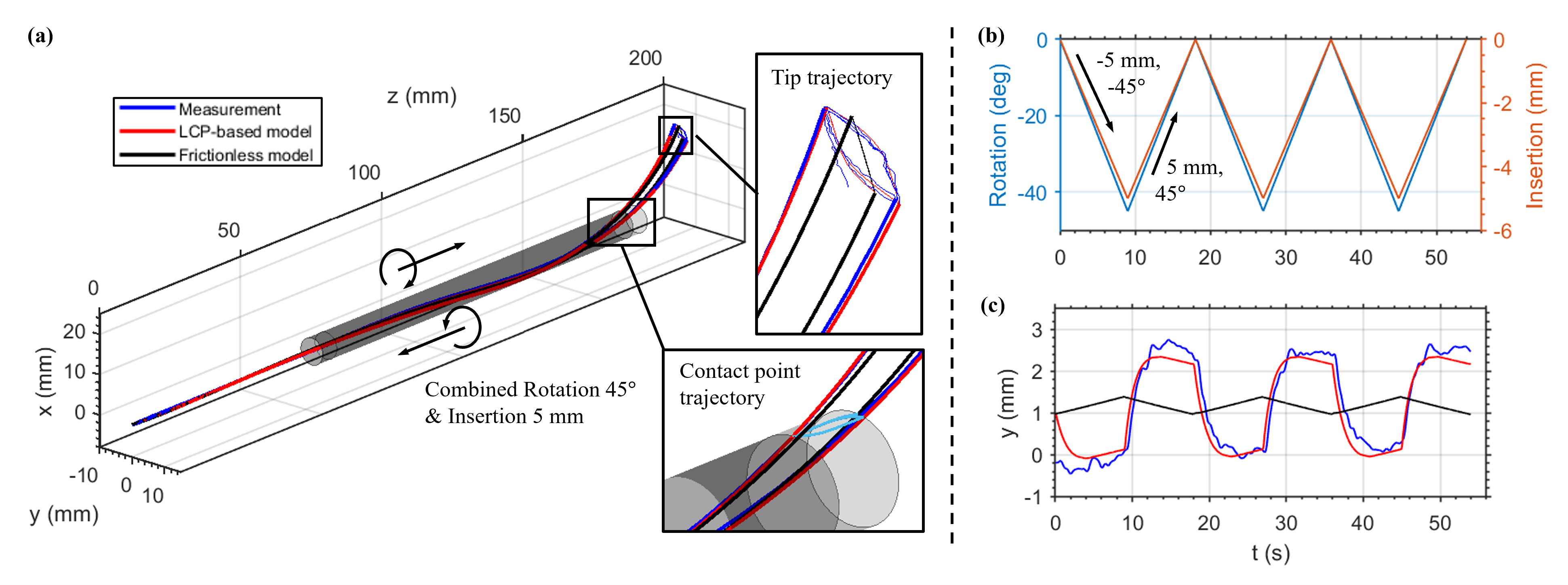}
    \caption{Representative experimental results compared with friction and frictionless model for the tube-pipe scenario when the pipe was rotated (45 degrees) and translated (5 mm) simultaneously and bidirectionally: (a) The 3D trajectory of the tube. The extreme shapes of the tube were shown when the direction of rotation/insertion shifted. The zoom-in plots at the tip and the contact region were presented. (b) The change of the actuation variables, including the rotation angle and insertion length of the pipe. (c) The change of the $y$-axis component of the tip position.}
    \label{fig:exp_cylinder_both}
\end{figure*}

Finally, we combined the bidirectional rotation and translation and studied the deflection of the tube under the frictional contact. The combined motion started at 2 different values of $h_c$, \{168.8, 178.8 mm\}, and the limits of the rotation and translation were 45$\degree$ and 5 mm, respectively. For each initial $h_c$, we examined the forward and backward direction of the first translation, for instance, $h_c$ w.r.t. forward translation varied from 168.8 mm to 173.8 mm. Four groups of experiments were conducted in total and the same COF $\mu_c=0.37$ was used.

The experiment with the initial $h_c= 178.8$ mm and a backward direction of the first translation was demonstrated in Fig. \ref{fig:exp_cylinder_both}. Fig. \ref{fig:exp_cylinder_both}(b) depicted the change of the actuation variables, namely the rotation angle and translation length. As shown in Fig. \ref{fig:exp_cylinder_both}(a), the tip trajectory was a closed curve consisting of a $z$-axis component due to translation and a $y$-axis component due to rotation. The proposed friction model accurately predicted the tip trajectory, while the frictionless model failed to capture the displacement along the $y$-axis. The change of the $y$-axis component of the tip position was shown in Fig. \ref{fig:exp_cylinder_both}(c). 

Note that the maximum tip deflection along the $y$-axis was reduced compared to the scenario of pure rotation in Fig. \ref{fig:exp_cylinder_rot}(c). For the same friction force magnitude, the added translation induced friction along the bending plane ($xz$-plane), leading to reduced friction force and less tip deflection along the $y$-axis. 
\corrlab{R1-3}{
This reduced out-of-plane deflection may also mitigate stick-slip motion, which is consistent with the lower value of maximum error metrics in Table~\ref{table_exp3_both} compared with Table~\ref{table_exp3_rot}. However, this interpretation cannot be independently verified from the current FBG measurements, since the corresponding error metrics are below the 1.11~mm mean body error reported for the FBG system.
}



\begin{table}[h!]
\renewcommand{\arraystretch}{1.25}
\caption{Error Analysis for the Tube-pipe Experiment (Combined Translation and Rotation) }
\label{table_exp3_both}
\small\sf\centering
\begin{tabularx}{0.36\textwidth}{c c p{0.07\linewidth} p{0.07\linewidth} p{0.07\linewidth}}
\toprule
Initial $h_c$ & Direction  & $\epsilon_{\text{tip}}$  & $\epsilon_{\text{body}}$ & $\epsilon_{\text{max}}$ \\
$[\text{mm}]$ &-- & $[\text{mm}]$  & $[\text{mm}]$ & $[\text{mm}]$ \\
\midrule 
168.8  &  Backward & 0.56 & 0.39 & 0.60 \\
168.8  &  Forward  & 0.63 & 0.41 & 0.79 \\
178.8  &  Backward & 0.35 & 0.38 & 0.73 \\
178.8  &  Forward  & 0.54 & 0.39 & 0.77 \\
\bottomrule
\end{tabularx}
\end{table}

\subsection{TDCR-plane Experiment} \label{sec:tdcr_plane}

\begin{figure}[h!]
    \centering
    \includegraphics[width = \linewidth]{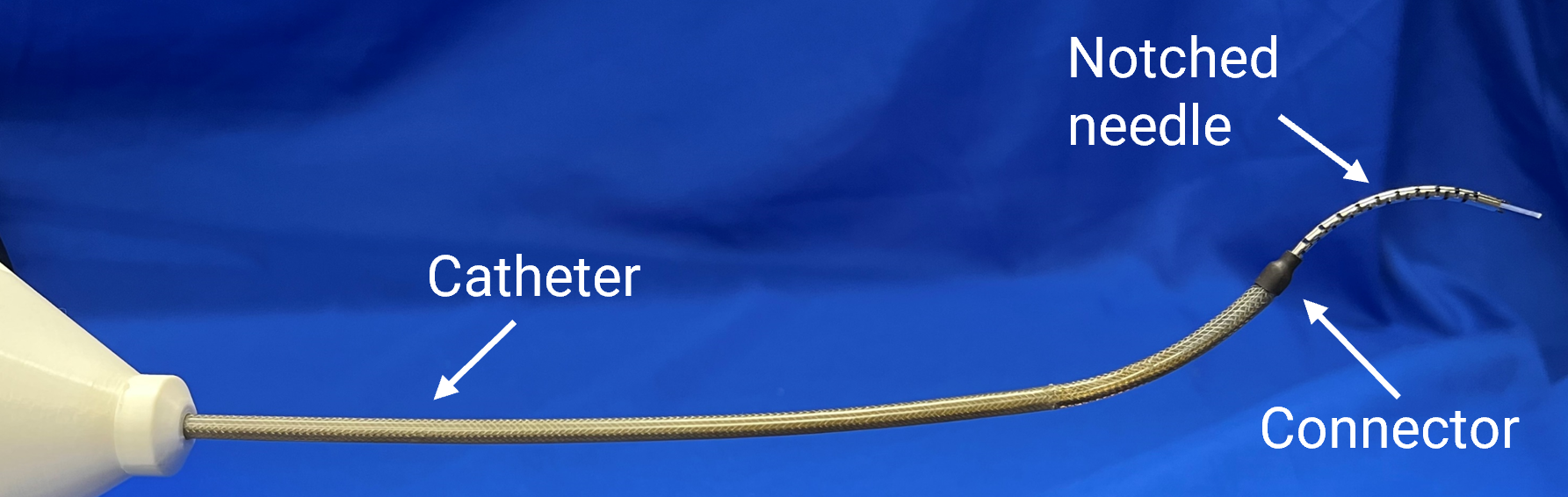}
    \caption{A tendon-driven continuum robot consisting of an outer catheter and a notched needle attached at the distal end. }
    \label{fig:exp_setup_tdcr}
\end{figure}

\begin{figure*}[t!]
    \centering
    \includegraphics[width = \linewidth]{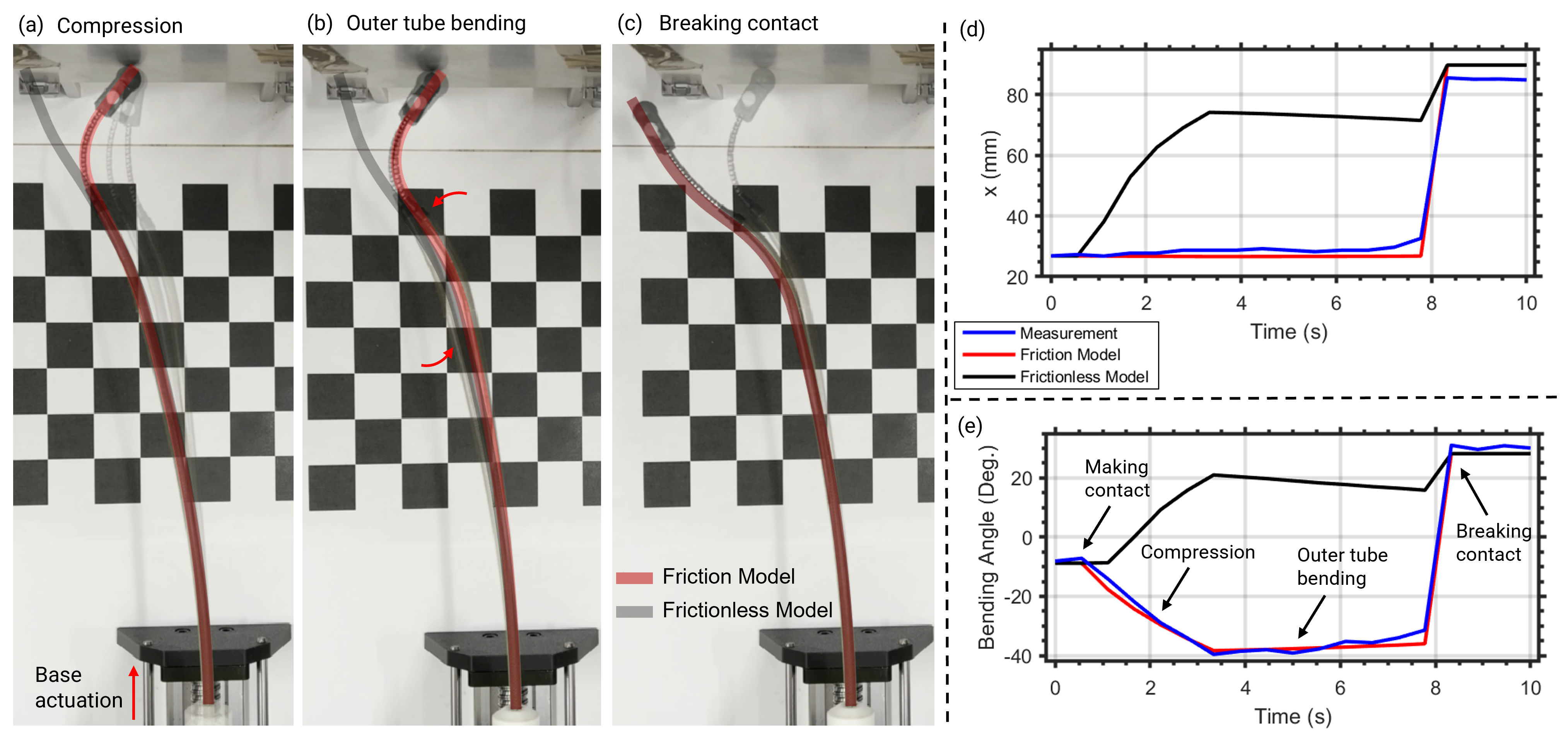}
    \caption{Experimental results of a tendon-driven continuum robot interacting with a rigid wall. The robot's motion sequence is shown with the path history rendered transparently, and model predictions overlaid. (a) The robot was compressed by translating the base, (b) Tip contact was maintained while the catheter was bent, and (c) Tip contact was broken as the catheter continued to bend. (d) Change in the $x$-component of the tip position. (e) Change in the tip's bending angle. 
    }
    \label{fig:exp_tdcr}
\end{figure*}

We further validated the proposed model on a TDCR composed of an outer catheter connected to a notched needle, as shown in Fig.~\ref{fig:exp_setup_tdcr}. Nitinol wires were used to actuate both the catheter and the notched needle, and were routed to constrain the robot’s motion inside a single plane. Base translation was also enabled to allow axial compression. Details on the robot's design and modeling will be presented separately in future work.

As shown in Fig.~\ref{fig:exp_tdcr}(a), we conducted an experiment of the TDCR interacting with a rigid wall (made of a 6061 aluminum sheet). A resin-printed part was connected to the tip of the notched needle to provide a stable frictional contact. The trajectory of the robot was captured via a camera (4032×3024  pixel resolution), and a checkerboard was used for camera calibration and distortion correction. The coefficient of friction (COF) was obtained by minimizing the mean trajectory error, resulting in a value of 0.092.

The robot was initially bent into an S-shape, and its base was actuated forward to establish contact with a rigid wall. In this stage, the robot was compressed by the contact force, but the tip remained stationary due to static friction. The catheter was then bent counterclockwise, as shown in Fig.~\ref{fig:exp_tdcr}(b), causing the robot to deflect while maintaining tip contact. As the friction reached the maximum, the static frictional contact broke, and the robot snapped to a new stable configuration (Fig.~\ref{fig:exp_tdcr}(c)) due to the sudden release of stored potential energy. The evolution of the tip's $x$-position and bending angles are shown in Fig.~\ref{fig:exp_tdcr}(d-e), respectively. The proposed model captured the smooth transition across these stages.

The maximum tip position error was 5.85 mm for the 287.9 mm long robot, and the maximum bending angle error was 4.6$\degree$, both occurring near the transition point where the contact with the wall was lost. 
These errors may be attributed to unmodeled stick-slip motions and internal friction induced by the tendons. In contrast, the frictionless model failed to capture the sticking motion of the tip, leading to a maximum tip position error of 45.4 mm and a bending angle error of 60.6$\degree$.

\section{Conclusion} \label{sec:conclusion}
This paper presents a novel external friction model of continuum robots based on complementarity constraints (CCs) that can accommodate static and sliding friction. The proposed method was applied to a case where an elastic tube made frictional contact with rigid environments. Experiments revealed that the maximum tip position error was reduced from $45.4$ mm in the frictionless model to $5.85$ mm for a $287.9$ mm-long robot. 

The main source of error is the unmodeled stick-slip motion when relative sliding between objects occurs. Modeling such an effect requires a more intricate friction model where the coefficient of friction ($\mu$) is time-varying and dependent on the relative sliding velocity \citep{fric_stickslip}. However, it results in a non-convex solution set and creates new difficulties \citep{Anitescu2006_convexCP}. 
Another limitation of the current model is that it relies on a single linearization step, so its accuracy depends on maintaining sufficiently small increments. This issue could potentially be mitigated through a more geometrically consistent update strategy, such as an inner iteration loop that repeatedly updates the contact state and re-linearizes the configuration within each step.
\corrlab{R1-1}{In addition, the formulation is quasi-static and neglects inertia, so it is not intended to capture highly dynamic effects such as snapping, impact, or very fast actuation, although such motions are often avoided in controlled continuum robot applications. The current contact formulation also assumes point contacts and therefore does not directly model distributed contact over a finite arc-length interval or continuous contact along curved surfaces. Finally, the experimental validation does not include scenarios dominated by elongation, shear, or pure torsional deformation of the continuum robot.}

While the proposed model focuses on the external friction between the robot and rigid environments, it paves the way toward accurate and efficient friction modeling in the continuum robot community. 
Our future work includes accommodating different robot structures, modeling environmental deformation, handling soft contacts, and extending the framework to strongly dynamic contact events, as briefly discussed below:
%
\begin{enumerate}
%
    \item Concentric-tube robots (CTRs). To model the internal tube-tube friction of CTRs, the deformation of both the inner and outer tubes due to friction should be considered. Meanwhile, the normal direction of the contact ($\mathbf{n}$) is assumed to be fixed over a time step, and the relative displacement ($\mathbf{v}$) is computed using the local linearization in (\ref{eq:pc_J_c}) and the relative-displacement relation in (\ref{eq:def_v1}). Since the tube radius of CTRs is small (less than 2 mm), a higher-order approximation of the surface may be necessary, which results in non-linear complementarity problems \citep{NonlinearCP}. 
    \item Soft contacts. Our model assumes hard contacts, i.e. concentrated point contact force. However, when the robot interacts with soft environments, for instance, needle steering in anatomical tissue or a phantom, a distributed load is applied. Therefore, the friction should be modeled as a distributed load \citep{Yang2022_DistributeFriction}, and the CCs (\ref{eq:CCs}) need to be reformulated.

    \item Dynamic effect. Our current model assumes quasi-static actuation of the continuum robot, which may be invalidated during abrupt stuck--jump--stuck transitions or other dynamic contact events. For mildly dynamic effects, these transitions may be alleviated by using finer incremental steps or introducing suitable regularization to improve robustness during abrupt contact-state changes \citep{TRO_simulator_review}. For strongly dynamic regimes, such as buckling, the present formulation is no longer sufficient and would require either a dynamic Cosserat rod contact model \citep{TRO2024_LCP} or a hybrid strategy that switches from the quasi-static LCP solver to a dynamic solver when abrupt transitions are detected.
        
\end{enumerate}

\begin{acks}
The authors would like to thank Chang Zhou for his help in developing the tendon-driven continuum robot.
\end{acks}


\begin{dci}
The author(s) declared no potential conflicts of interest with respect to the research, authorship and/or publication of this article.
\end{dci}

\begin{funding}
Research reported in this publication was supported by the Georgia Tech Faculty Startup Grant, NSF Awards 2339202 and 2514173, and USDA-NIFA’s National Robotics Initiative Award 2022-11065 in collaboration with the National Science Foundation. It was also partially supported by the National Institute of Biomedical Imaging and Bioengineering of the National Institutes of Health under Award Number R01EB034359, and the National Heart Lung and Blood Institute of the National Institutes of Health under Award Number R01HL183894. The content is solely the responsibility of the authors and does not necessarily represent the official views of the National Institutes of Health. Dr. Junhyoung Ha is also partially supported by the National Research Foundation (NRF) of Korea under the grant No. RS-2022-NR071784 funded by the Korea government (MSIT).
\end{funding}


\bibliographystyle{SageH} 
\bibliography{reference.bib} 

\begin{appendices}
\setcounter{secnumdepth}{1}

\section{Index to multimedia extensions} 
\label{app:multimedia}
\setcounter{table}{0}
\renewcommand{\thetable}{\thesection\arabic{table}}
\renewcommand{\theHtable}{appendix.\thesection.\arabic{table}}

\begin{table}[h!]
\caption{Table of multimedia extensions}
\renewcommand{\arraystretch}{1.25}
\label{appendix}
\small\sf\centering
\begin{tabularx}{0.5\textwidth}{l l X}
\toprule
Extension  & Media type  & Description \\
\midrule 
1  &         Video       & {Animations of LCP-based model simulations and experimental validations.} \\
\bottomrule
\end{tabularx}
\end{table}

\end{appendices}

\end{document}